\documentclass{bmvc2k}
\usepackage{subcaption}
\usepackage{xcolor}
\usepackage{colortbl}
\usepackage{enumitem}
\usepackage{multirow}
\usepackage{algorithm}
\usepackage{algpseudocode}
\usepackage{amsmath,amssymb}
\usepackage{wrapfig}
\usepackage{graphicx}
\usepackage{amssymb,amsfonts}
\usepackage{booktabs}

\title{
\includegraphics[height=1.8ex]{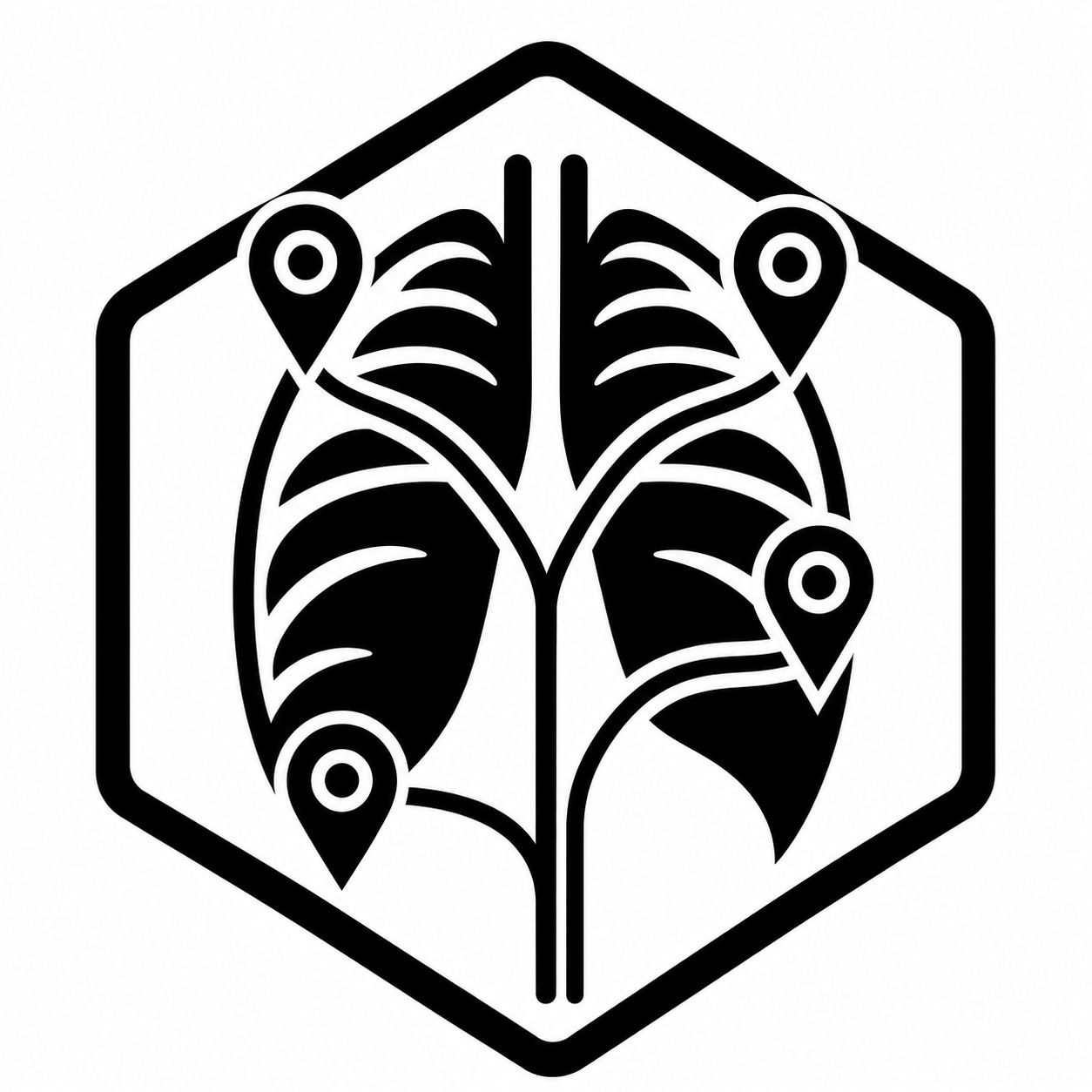}CheXGround: Anatomical Region Tokens for Grounded Longitudinal Chest X-ray Interpretation
}

\addauthor{Adonay Demewez Gebremedhin}{202311465@ajmanuni.ac.ae}{1}
\addauthor{Wessam Shehieb}{w.shehieb@ajman.ac.ae}{1}
\addauthor{Sara Alansari}{saraansari@live.com}{2}
\addauthor{Mohamad Alansari}{100061914@ku.ac.ae}{3}
\addauthor{Muzammal Naseer}{muhammadmuzammal.naseer@ku.ac.ae}{3,4}
\addauthor{Sajid Javed}{sajid.javed@ku.ac.ae}{3}
\addauthor{Naoufel Werghi}{Naoufel.Werghi@ku.ac.ae}{3}

\addinstitution{
Ajman University\\
Ajman, United Arab Emirates
}

\addinstitution{
University of Birmingham\\
Birmingham, United Kingdom
}

\addinstitution{
Khalifa University\\
Abu Dhabi, United Arab Emirates
}

\addinstitution{
University of Western Australia\\
Australia
}

\runninghead{A. D. Gebremedhin et al.}{CheXGround}

\newcommand{\gain}[1]{\,{\tiny\textcolor{black!55}{[\ensuremath{\uparrow}#1]}}}

\begin{document}

\maketitle
\begin{abstract}
\noindent Recent radiology multi-modal language models have made substantial progress in chest X-ray report generation, visual question answering, and temporal reasoning. 
While longitudinal chest X-ray interpretation compares sequential examinations to describe change, visual grounding aims to connect clinical language with localized image evidence.
Although longitudinal modeling and visual grounding have each advanced radiology language models, how localized visual evidence can support longitudinal interpretation remains under-explored.
We introduce CheXGround, a region-grounded longitudinal chest X-ray language model that represents paired studies through corresponding anatomical regions. 
CheXGround extracts anatomical regions from current and prior radiographs, encodes them as temporally enhanced Region-of-Interest (ROI) tokens, and combines them with global temporal image context during generation. 
To connect these region tokens with clinical text, we propose Temporal Region--Phrase Alignment, a pretraining objective that aligns temporal anatomical representations with localized report phrases. 
We evaluate CheXGround on single-study and longitudinal Visual Question Answering (VQA), longitudinal findings generation, temporal grounded VQA, and anatomical grounding. 
Across these tasks, CheXGround improves clinical language quality, temporal reasoning, and localization accuracy over recent baselines. 
Our results suggest that organizing longitudinal evidence at the anatomical level is a strong representation for grounded radiology language modeling. \\ Project Page: \url{https://adonaydem.github.io/chexground-website}
\end{abstract}

%-------------------------------------------------------------------------
\section{Introduction}
\label{sec:intro}
\noindent Radiologists often rely on structured diagnostic evaluation, relating fine-grained spatial cues to abnormalities, comparing findings across anatomical regions to test clinical hypotheses, and incorporating patient history to assess disease progression~\cite{Klein2019,abbott2019radiologyreport}. 
These single and longitudinal workflows require models to represent both localized anatomy and temporal change. 
In longitudinal scenarios, the clinically important signal is often not a global difference between two images, but a subtle change in a specific region, such as a worsening \emph{basilar} opacity, a resolving \emph{right lower-lobe} consolidation, or location of a newly visible support device. 
A model for longitudinal chest X-ray interpretation should therefore connect \emph{what} finding is present, \emph{where} it is located, and \emph{how} it has changed over time.

\begin{figure}[t]
    \centering
    \includegraphics[
        width=\textwidth,
    ]{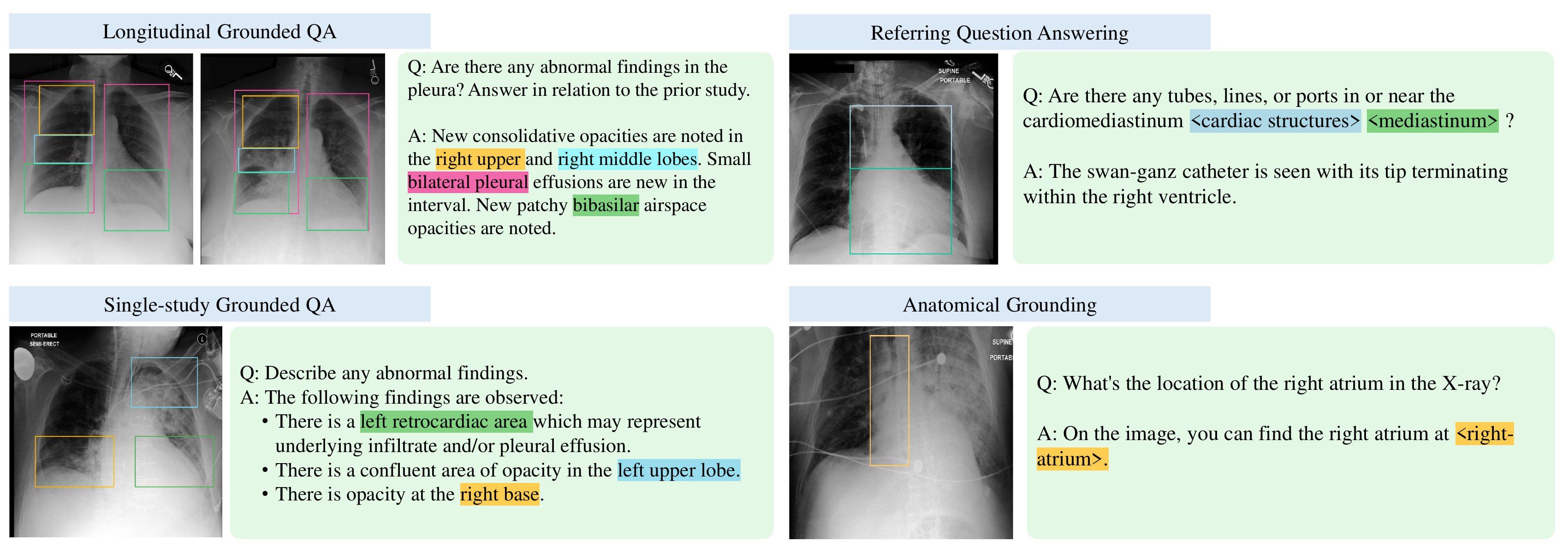}
    \captionsetup{skip=6pt}
    \caption{
\textbf{Overview of the grounded radiology tasks studied in this work.}
CheXGround supports grounded interaction for chest X-ray interpretation, spanning spatio-temporal comparison, finding-level reasoning, referring conversations, and anatomical localization. 
These tasks move beyond answer generation alone by requiring radiology reasoning to be grounded across multiple granularities.
}
\label{fig:task_hook}
\end{figure}

General-domain grounding Multi-modal Large Language Models (MLLMs) have shown that generated language can be effectively tied to visual evidence through bounding boxes, segmentation masks, and spatio-temporal grounding~\cite{lai2024lisa,bai2024one,rasheed2024glamm,ma2024groma,qiu2024artemis,munasinghe2025videoglamm,wang2026spacevllm}. 
In medicine, recent developments improve local image--text alignment through localized contrastive learning, phrase grounding, region-level representation learning, multi-granularity supervision, and anatomy-aware pretraining~\cite{huang2021gloria,muller2022joint,wang2022multi,rizvi2024local,li2024anatomical,zhang2025anatomical,muller2024chexinteractivelocalizationregion}. Recent grounded medical MLLMs further move this direction toward localized generation by linking generated answer text to visual evidence such as boxes or masks~\cite{bannur2024maira,luo2025vividmed, zhou2024medversa,chen2024chexagent, deperrois2025radvlm, choi2026instructionguidedlesionsegmentationchest}. 
Longitudinal and difference-aware chest X-ray models make complementary progress by incorporating prior examinations for temporal report generation, progression modeling, difference visual question answering, and temporal vision--language pretraining~\cite{karwande2022chexrelnet,zhu2023utilizing,hu2023medicaldiff,cho2024pretraining,lu2024spot,yang2025tempa,wang2024hergen,song2025ddatr,zhang2025libra}. 
However, tightly coupling spatial grounding with longitudinal comparison in radiology remains open, as clinically meaningful change may depend on how localized evidence evolves across prior and current studies.

Turning this motivation into a grounded longitudinal radiology model is non-trivial for four reasons.
\emph{(i)} Grounded radiology MLLMs often separate question answering from box or coordinate prediction rather than composing both simultaneously~\cite{chen2024chexagent,deperrois2025radvlm,zhou2024medversa}.
We instead use dual vision--language interleaving, composing clinical text with localized visual evidence within the same reasoning and generation process.
\emph{(ii)} Clinical interaction is multi-granular, requiring responses that connect findings, anatomical references, and supporting regions~\cite{muller2026a}.
We therefore use dense grounded supervision interleaving phrase-level visual annotations with clinical answers.
\emph{(iii)} Some methods localize through coordinate regression in the language output~\cite{deperrois2025radvlm,chen2024chexagent}, while others ground after generation~\cite{zhou2024medversa}, leaving visual evidence weakly integrated into reasoning~\cite{ma2024groma}.
We instead expose anatomical regions as visual tokens, enabling language generation over region-level evidence compared across current and prior studies.
\emph{(iv)} Grounded longitudinal modeling requires large-scale supervision of dense spatio-temporal relations.
We therefore train on 2.2M samples spanning language generation, visual grounding, and temporal interpretation~\cite{muller2026a,bae2023ehrxqa,hu2023medicaldiff,deperrois2025radvlm}.
Together, these challenges motivate reasoning over anatomical evidence across current and prior studies.

We introduce \textbf{CheXGround}, a \textbf{Che}st \textbf{X}-ray \textbf{Ground}ed longitudinal MLLM for region-referential chest X-ray interpretation.
CheXGround represents each study using a fixed set of anatomical regions, compares each region in the current radiograph with the same region in the prior radiograph, and passes the resulting ROI tokens to the language model.
Because these tokens correspond to fixed chest anatomy, the model can reason over clinically meaningful regions rather than only global image tokens.
This follows the broader idea of localized visual tokenization~\cite{ma2024groma}, but adapts it to radiology, where the relevant regions are anatomical structures shared across exams.
To connect these regions with clinical language, we pretrain them with Temporal Region--Phrase Alignment (TRPA), which combines an ROI-level contrastive objective inspired by GLoRIA~\cite{huang2021gloria} with phrase-composition targets that allow related clinical phrases to share localized evidence.
CheXGround then uses both global temporal image tokens and phrase-aligned anatomical ROI tokens for grounded visual question answering, disease-progression reasoning, and longitudinal report generation (See Fig.~\ref{fig:task_hook}).
\noindent Our contributions are:
\begin{itemize}[leftmargin=1.25em, itemsep=2pt, topsep=2pt]
    \item We introduce \emph{CheXGround}, a two-stream longitudinal MLLM using global-temporal image tokens and temporal anatomical ROI tokens.

    \item We propose Temporal Region--Phrase Alignment pretraining (TRPA), with \emph{GLoRIA-ROI} and phrase-composition alignment for temporally grounded ROI representations.

  \item We evaluate CheXGround on diverse visual grounding tasks demonstrating consistent gains over baselines in both clinical language quality and grounding accuracy.
\end{itemize}

\section{Related Work}
\label{sec:related}

\noindent\textbf{Radiology vision-language models.}
Report generation has progressed from CNN, recurrent, transformer, hierarchical, co-attention, and memory-based models to methods using medical knowledge, abnormality graphs, and retrieval priors~\cite{jing2018automatic,wang2018tienet,chen2020generating,chen2021cross,li2019knowledge,liu2021exploring}.
Medical and difference-aware VQA extend this to question-conditioned reasoning over abnormalities, locations, evidence, and temporal change~\cite{lau2018dataset,ben2019vqa,liu2021slake,he2020pathvqa,kornuta2019leveraging,li2023self,zhan2025uniclam,gai2024medthink,bae2023ehrxqa,nguyen2025vindr,hu2023medicaldiff,cho2024pretraining,lu2024spot}.
Contrastive pretraining improves image--text representations through large-scale image--report alignment~\cite{zhang2022contrastivelearningmedicalvisual,boecking2022making,bannur2023learning}, while MLLMs support conversational reasoning, reporting, and VQA~\cite{li2023llava,chen2024chexagent,bannur2024maira,deperrois2025radvlm,park2025m4cxr,zhou2024medversa,zhang2025libra,sellergren2026medgemma,hyland2023maira}.
These advances provide strong vision-language backbones, while our work explicitly grounds reasoning in anatomical evidence for single-study and longitudinal interpretation.

\noindent\textbf{Grounded medical and chest X-ray understanding.}
Grounded medical vision-language learning links clinical text to visual evidence through localized contrastive learning, sentence--region alignment, multi-granularity supervision, anatomy-aware representations, and interactive localization~\cite{huang2021gloria,muller2022joint,wang2022multi,rizvi2024local,li2024anatomical,zhang2025anatomical,muller2024chexinteractivelocalizationregion}.
General grounding MLLMs extend this to segmentation, region references, pixel-level and spatial reasoning, and spatio-temporal grounding~\cite{lai2024lisa,bai2024one,rasheed2024glamm,munasinghe2025videoglamm,alansari2026sparrowlearningspatialprecision,ma2024groma,qiu2024artemis,wang2026spacevllm}.
Radiology-specific methods support grounded reporting and localization with boxes, masks, and multi-modal inputs~\cite{bannur2024maira,deperrois2025radvlm,luo2025vividmed,zhou2024medversa}, while datasets provide anatomy-centered scene graphs, abnormality boxes, and grounded VQA~\cite{wu2021chest,nguyen2025vindr,de_Castro_2025,muller2026a}.
These works connect clinical text with localized evidence, but remain largely non-temporal. Our work instead models anatomical evidence across current and prior studies for fine-grained spatio-temporal grounding.

\noindent\textbf{Longitudinal chest X-ray modeling.}
Longitudinal chest X-ray interpretation uses prior studies to assess whether findings are new, improving, worsening, stable, or resolved.
Task-specific models incorporate priors through image/report conditioning, cross-attention, history-aware or region-specific generation, difference modeling, residual alignment, and temporal pretraining~\cite{karwande2022chexrelnet,zhu2023utilizing,dalla2023controllable,wang2024hergen,song2025ddatr,hu2023medicaldiff,cho2024pretraining,lu2024spot,bannur2023learning,yang2025tempa}.
Recent radiology MLLMs extend this to instruction following, multi-task temporal understanding, and generalist interpretation~\cite{zhang2025libra,zhou2024medversa,chen2024chexagent,sellergren2026medgemma,bannur2024maira}.
These works show the value of priors for temporal understanding. CheXGround extends this toward anatomy-aware temporal modeling, representing current--prior changes through localized anatomical evidence for fine-grained grounded reasoning.

%-------------------------------------------------------------------------

\section{Methodology}
\label{sec:method}

\subsection{Overview and Problem Formulation}
\label{sec:method_overview}

\noindent \textbf{Problem formulation and architecture}. 
Given radiographs $\{I_t\}_{t=1}^{T}$ and an instruction $x$, CheXGround generates a response for spatial or spatio-temporal VQA tasks.
Even though our method can be applied to arbitrary $T$, we focus on the common paired setting $(I_{\mathrm{curr}}, I_{\mathrm{prior}})$ with $T=2$ following recent literature. 
The model combines two visual streams: global temporal image tokens for holistic context, and anatomical region tokens for localized evidence from fixed chest regions. 
Building on region-referential grounded MLLMs~\cite{lai2024lisa,bai2024one,rasheed2024glamm,ma2024groma,qiu2024artemis,wang2026spacevllm}, CheXGround instantiates $K$ anatomical slots from Chest ImaGenome~\cite{wu2021chest}, localizes each slot in both studies, and encodes it as an ROI embedding $\mathbf{v}^{\mathrm{roi}}_k$ that carries anatomical identity and localized temporal context. These embeddings are inserted into the language model through a compact region block:
\begin{equation}
[
\langle r_1\rangle \langle \mathrm{region}\rangle_1,
\ldots,
\langle r_K\rangle \langle \mathrm{region}\rangle_K
],
\end{equation}
where $\langle r_k\rangle$ denotes the $k$-th anatomical slot and the embedding of $\langle \mathrm{region}\rangle_k$ is replaced by $\mathbf{v}^{\mathrm{roi}}_k$. 
These region token handles allow both model outputs and user instructions to refer to explicit anatomical evidence~\cite{ma2024groma}.

\subsection{Temporal Anatomical Region-of-Interest Tokens}
\label{sec:region_tokens}

%=============================================================
\begin{figure*}[t]
  \centering
  \begin{minipage}[b]{0.65\textwidth}
    \centering
    \includegraphics[width=\linewidth]{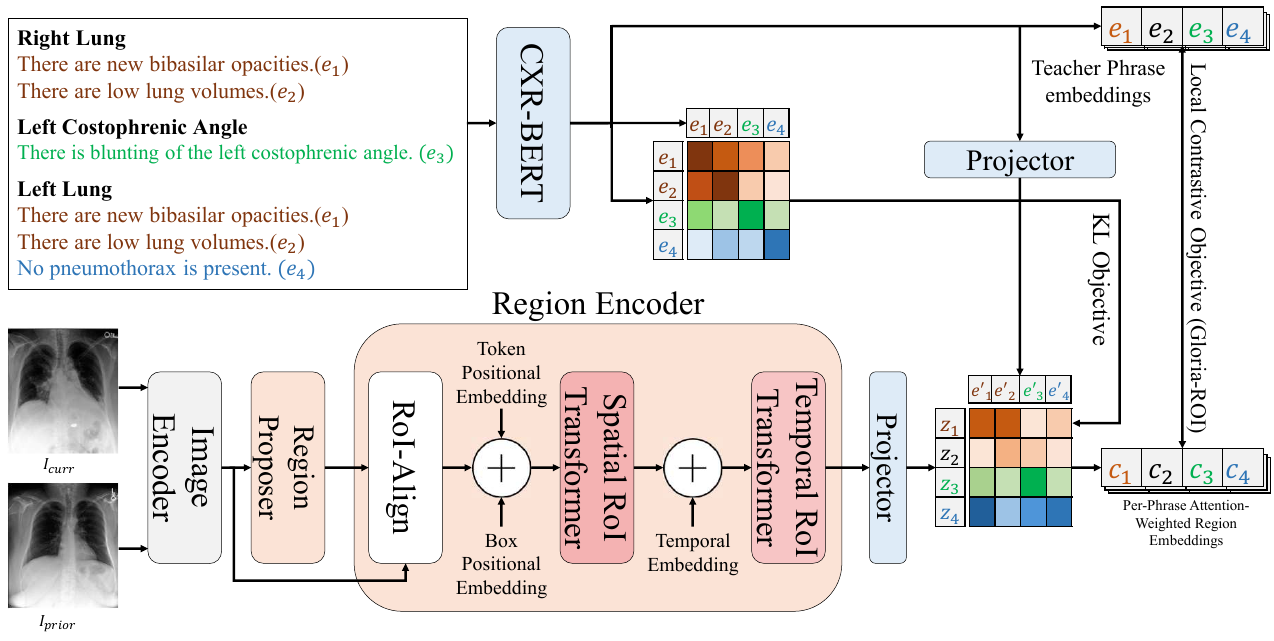}
    \textbf{(a)}
  \end{minipage}
  \hfill
  \begin{minipage}[b]{0.15\textwidth}
    \centering
    \includegraphics[width=\linewidth]{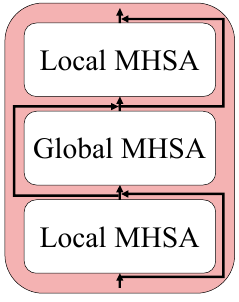}
    \textbf{(b)}
  \end{minipage}
  \hfill
  \begin{minipage}[b]{0.15\textwidth}
    \centering
    \includegraphics[width=\linewidth]{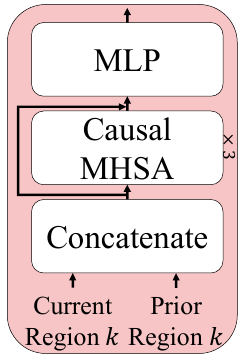}
    \textbf{(c)}
  \end{minipage}
  \vspace{0.5em}
\caption{
\textbf{Temporal anatomical ROI encoding and region--phrase alignment.}\\
(a) TRPA encodes current and prior chest X-rays, extracts anatomical regions with ROIAlign~\cite{he2017mask}, and aligns them with localized report phrases. 
The spatial ROI transformer in (b) refines each region using local and global context, while the temporal ROI transformer in (c) models current and prior causal relationships. \emph{MHSA denotes multi-head self-attention.}
}
  \label{fig:temporal_anatomical}
\end{figure*}
% =============================================================

We represent localized longitudinal evidence with fixed anatomical slots corresponding to clinically meaningful chest regions.
For each slot $k$, CheXGround preserves anatomical identity while encoding its current appearance and change relative to the prior study.

\noindent\textbf{Anatomical ROI extraction.}
For each radiograph $I_t$, we extract multiscale pyramid features
$\{\mathbf{F}^{\ell}_{t}\}_{\ell=1}^{L}
=\mathrm{Pyr}(\phi(I_t))$
using the RAD-DINO image backbone $\phi$~\cite{perezgarcia2024raddino}.
An anatomy-aware detector $\mathcal{D}$, based on Deformable DETR~\cite{zhu2021deformable}, predicts one box per anatomical slot:
\begin{equation}
\{\mathbf{b}_{t,k}\}_{k=1}^{K}
=
\mathcal{D}(\{\mathbf{F}^{\ell}_{t}\}_{\ell=1}^{L}),
\qquad
\mathbf{b}_{t,k}\in[0,1]^4 .
\end{equation}
The detector is pretrained from scratch on Chest ImaGenome~\cite{wu2021chest} box annotations and used in subsequent tasks.
Given $\mathbf{b}_{t,k}$, ROIAlign~\cite{he2017mask} extracts an $A\times A$ local feature grid:
\begin{equation}
\mathbf{U}_{t,k}
=
\mathrm{ROIAlign}(
\{\mathbf{F}^{\ell}_{t}\}_{\ell=1}^{L},
\mathbf{b}_{t,k})
\in \mathbb{R}^{A^2\times d}.
\end{equation}

\noindent\textbf{ROI transformer block.}
We use the following transformer block for spatial and temporal ROI encoding:
\begin{equation}
\begin{aligned}
\mathbf{Y} &= \mathbf{X}+\mathrm{MHSA}(\mathrm{LN}(\mathbf{X})),\\
\mathcal{B}(\mathbf{X}) &= \mathbf{Y}+\mathrm{MLP}(\mathrm{LN}(\mathbf{Y})),
\end{aligned}
\label{eq:roi_transformer_block}
\end{equation}
for input tokens $\mathbf{X}\in\mathbb{R}^{n\times d}$, where $\mathrm{LN}$, $\mathrm{MHSA}$, and $\mathrm{MLP}$ denote layer normalization, multi-head self-attention~\cite{vaswani2017attention}, and a multi-layer perceptron.

\noindent\textbf{Spatial ROI encoding.}
Because cropped ROIs can lose surrounding anatomical context~\cite{chen2020hierarchical}, each grid $\mathbf{U}_{t,k}$ is augmented with a learnable box-coordinate embedding and two-dimensional sinusoidal grid embeddings, yielding
$\widetilde{\mathbf{U}}_{t,k}\in\mathbb{R}^{N\times d}$, where $N=A^2$.
As shown in Fig.~\ref{fig:temporal_anatomical}(b), the spatial ROI transformer is
\begin{equation}
\begin{aligned}
\mathbf{L}_{t,k} &= \mathcal{B}_{\ell 1}(\widetilde{\mathbf{U}}_{t,k}),\\
[\mathbf{G}_{t,1};\ldots;\mathbf{G}_{t,K}]
&= \mathcal{B}_{g}([\mathbf{L}_{t,1};\ldots;\mathbf{L}_{t,K}]),\\
\mathbf{H}_{t,k} &= \mathcal{B}_{\ell 2}(\mathbf{G}_{t,k}),
\end{aligned}
\end{equation}
where $[\cdot;\cdot]$ denotes token concatenation and the global output is split back into anatomical slots.
The first local block $\mathcal{B}_{\ell 1}$ models the $N$ tokens within each ROI, $\mathcal{B}_{g}$ attends over all $KN$ image tokens to recover cross-anatomical context, and $\mathcal{B}_{\ell 2}$ re-integrates this context within each ROI.

\noindent\textbf{Causal temporal encoding.}
To model interval change without mixing unrelated anatomy, temporal attention is applied independently within each slot.
The ordered spatial grids are augmented with 1-D sinusoidal temporal embeddings, yielding $\widetilde{\mathbf{H}}_{k}$.
As shown in Fig.~\ref{fig:temporal_anatomical}(c), the causal temporal ROI transformer applies the same block $\mathcal{B}$ with a causal mask $\mathbf{M}_{c}$ along the study axis at each ROI-grid location:
\begin{equation}
\{\mathbf{Z}_{t,k}\}_{t=1}^{T}
=
\mathcal{E}_{t}^{\mathrm{causal}}
(\widetilde{\mathbf{H}}_{k};\mathbf{M}_{c}),
\qquad
\mathbf{Z}_{t,k}\in\mathbb{R}^{N\times d},
\end{equation}
where $\mathcal{E}_{t}^{\mathrm{causal}}$ denotes a stack of blocks $\mathcal{B}$ followed by a multi-layer perceptron.
In the paired setting, current ROI tokens condition on their prior counterparts, whereas prior tokens cannot access current evidence~\cite{wang2024hergen}, enabling anatomically matched temporal comparison.

%-------------------------------------------------------------------------

\subsubsection{Temporal Region--Phrase Alignment Pretraining}
\label{sec:trpa}

The temporal region encoder produces anatomically structured ROI tokens, but anatomical structure alone does not ensure that these tokens form clinically meaningful representations.
We therefore introduce Temporal Region--Phrase Alignment pretraining (TRPA), which aligns temporally contextualized anatomical ROIs with localized report phrases before they are passed to the language model in Section~\ref{sec:CheXGround_lm}. 
For each slot, we average-pool the current ROI grid as $\mathbf{z}_{k}=\mathrm{AvgPool}(\mathbf{Z}_{T,k})\in\mathbb{R}^{d}$, where $T$ denotes the current (most recent) study. 
From the associated report, we extract clinically relevant phrase spans $\mathcal{P}_{i}$, including findings and devices with potential temporal-change expressions. 
Each phrase $p\in\mathcal{P}_{i}$ is encoded by a text encoder $\psi(\cdot)$ and projected non-linearly into the ROI feature space as:
\begin{equation}
    \mathbf{e}'_{p}=g_{\mathrm{text}}(\psi(p))\in\mathbb{R}^{d}.
\end{equation}

\noindent \textbf{ROI-level spatio-temporal alignment.}
Radiology reports describe findings through phrases or statements that often point to specific anatomical regions. 
TRPA therefore aligns each phrase with its supporting anatomical ROI.
For a candidate image--report pair $(I,R)$ and phrase $p\in\mathcal{P}$, we compute its similarity to each anatomical slot $k$:
\begin{equation}
    s_{p,k}
    =
    (\mathbf{e}'_{p})^{\top}\mathbf{z}_{k}.
\end{equation}
The phrase then induces a soft attention distribution over anatomical ROIs:
\begin{equation}
    a_{p,k}
    =
    \frac{
        \exp(s_{p,k}/\tau_{a})
    }{
        \sum_{k'=1}^{K}\exp(s_{p,k'}/\tau_{a})
    },
\end{equation}
where $\tau_a$ controls the sharpness of phrase-to-region attention. 
Using these attention weights, we form a phrase-conditioned visual representation:
\begin{equation}
    \mathbf{c}_{p}
    =
    \sum_{k=1}^{K}
    a_{p,k}\mathbf{z}_{k}.
\end{equation}

The compatibility between phrase $p$ and the image is measured by comparing the phrase representation with its attended visual context. 
We aggregate phrase-level compatibilities into an image--report similarity score:
\begin{equation}
    S(I,R)
    =
    \tau_{p}
    \log
    \left(
    \frac{1}{|\mathcal{P}|}
    \sum_{p\in\mathcal{P}}
    \exp
    \left(
        \frac{
            \mathbf{c}_{p}^{\top}\mathbf{e}'_{p}
        }{\tau_{p}}
    \right)
    \right),
\end{equation}
where $\tau_p$ controls the smoothness of phrase aggregation. 
For a minibatch of $B$ matched image--report pairs, we apply this scoring function to all candidate pairs and denote the score between image $i$ and report $j$ as $S_{ij}=S(I_i,R_j)$. 
The ROI-level local contrastive objective is:
\begin{equation}
    \mathcal{L}_{\mathrm{roi}}^{\mathrm{img}\rightarrow\mathrm{text}}
    =
    -\frac{1}{B}
    \sum_{i=1}^{B}
    \log
    \frac{
        \exp(S_{ii}/\tau_{\mathrm{loc}})
    }{
        \sum_{j=1}^{B}\exp(S_{ij}/\tau_{\mathrm{loc}})
    },
\end{equation}
\begin{equation}
    \mathcal{L}_{\mathrm{roi}}^{\mathrm{text}\rightarrow\mathrm{img}}
    =
    -\frac{1}{B}
    \sum_{j=1}^{B}
    \log
    \frac{
        \exp(S_{jj}/\tau_{\mathrm{loc}})
    }{
        \sum_{i=1}^{B}\exp(S_{ij}/\tau_{\mathrm{loc}})
    },
\end{equation}
where $\tau_{\mathrm{loc}}$ is the contrastive temperature. 
We define the symmetric GLoRIA-ROI objective as:
\begin{equation}
    \mathcal{L}_{\mathrm{GLoRIA\text{-}ROI}}
    =
    \frac{1}{2}
    \mathcal{L}_{\mathrm{roi}}^{\mathrm{img}\rightarrow\mathrm{text}}
    +
    \frac{1}{2}
    \mathcal{L}_{\mathrm{roi}}^{\mathrm{text}\rightarrow\mathrm{img}}.
\end{equation}
Li et al.~\cite{li2024anatomical} construct anatomy-aware region--sentence positives from single-study boxes and parsed reports. 
In contrast to this single-study sentence-level supervision, and to patch-based global-local objectives such as GLoRIA~\cite{huang2021gloria} and BioViL-T~\cite{bannur2023learning}, our alignment lets phrase evidence match temporally contextualized region semantics rather than single-study anatomical assignments or noisy dense patches, as supported by Table~\ref{tab:trpa_retrieval_ablation}.

\noindent \textbf{Phrase-composition targets.}
ROI-level contrastive alignment encourages phrase tokens to strictly attend to supporting regions, but radiology language is sparse and compositional, so related phrases can share overlapping anatomical evidence. Moreover, because $\mathcal{L}_{\mathrm{GLoRIA\text{-}ROI}}$ is ultimately aggregated at the image--report level, it provides limited fine-grained region semantics. We therefore use phrase-composition targets that allow related clinical phrases to share evidence, rather than forcing each phrase to match a single region exclusively.

\begin{wrapfigure}{r}{0.58\textwidth}
    \centering
    \includegraphics[width=\linewidth]{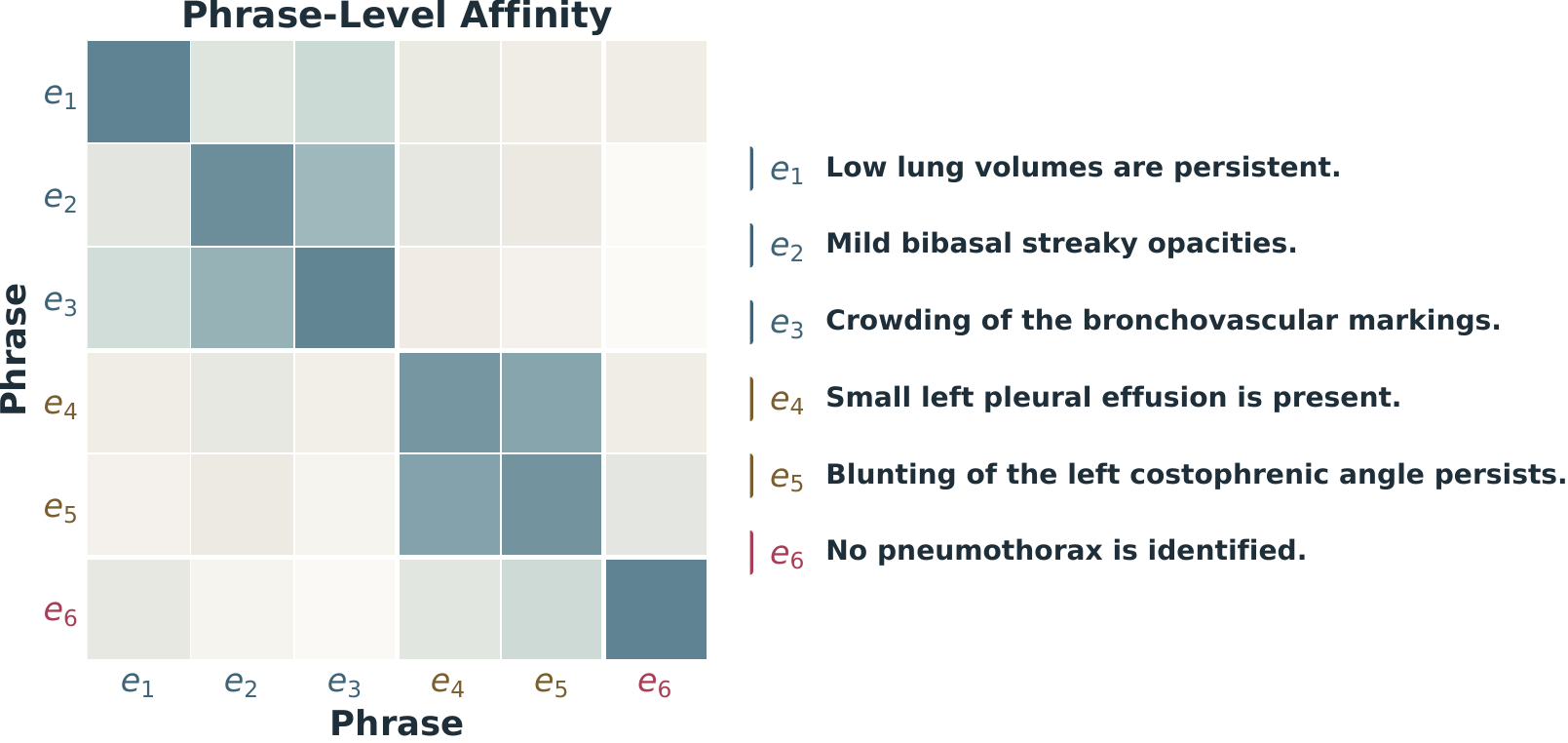}
    \captionsetup{font=small,skip=3pt}
    \caption{
    Phrase-level affinity computed via CXR-BERT~\cite{bannur2023learning} captures spatio-temporal semantic relationships between diverse region findings. Darker cells indicate higher pair similarity.
    }
    \label{fig:phrase_affinity}
\end{wrapfigure}

For each annotated image and anatomical slot $k$, we define a hard target distribution
$q^{\mathrm{hard}}_{k}\in\Delta^{|\mathcal{P}|}$ over the phrase set $\mathcal{P}$ using Chest ImaGenome~\cite{wu2021chest}.
This distribution assigns probability mass to the phrases linked to slot $k$, while neglecting the rest. 

However, this hard assignment treats all unannotated phrases as equally incorrect, even when some phrases are semantically close to the annotated ones and may describe similar visual evidence. 
We therefore relax the hard target by spreading a portion of its mass to related phrases, while keeping the original localized annotation as the primary supervision signal. To account for semantic similarity among phrases, we construct a phrase-level affinity matrix using the original frozen text embeddings:
\begin{equation}
    A_{p,p'}
    =
    \frac{
        \exp\left(\mathbf{e}_{p}^{\top}\mathbf{e}_{p'} / \tau_{\mathrm{a}}\right)
    }{
        \sum_{\bar{p}\in\mathcal{P}}
        \exp\left(\mathbf{e}_{p}^{\top}\mathbf{e}_{\bar{p}} / \tau_{\mathrm{a}}\right)
    },
\end{equation}
where $\mathbf{e}_{p}$ denotes the original frozen text embedding of phrase $p$, and $\tau_{a}$ controls the sharpness. Figure~\ref{fig:phrase_affinity} demonstrates the potential of our phrase-level affinity setup.

\noindent The final softened target for slot $k$ in image $i$ is then defined as:
\begin{equation}
q_{i,k,p}
=
(1-\alpha)q^{\mathrm{hard}}_{i,k,p}
+
\alpha
\sum_{a\in\mathcal{P}}
q^{\mathrm{hard}}_{i,k,a}A_{a,p},
\end{equation}
where $\alpha\in[0,1]$ controls soft targets' importance. The first term retains the original hard target for phrase $p$, scaled by $(1-\alpha)$. The second term accumulates mass from all phrases $a\in\mathcal{P}$, weighted by their semantic affinity $A_{a,p}$ to phrase $p$.

The model predicts an ROI-to-phrase distribution by comparing the pooled anatomical representation for slot $k$ in image $i$ with each projected phrase embedding:
\begin{equation}
\hat{q}_{i,k,p}
=
\frac{
\exp\left(\mathbf{z}_{i,k}^{\top}\mathbf{e}'_{p}/\tau_{\mathrm{ph}}\right)
}{
\sum_{p'\in\mathcal{P}}
\exp\left(\mathbf{z}_{i,k}^{\top}\mathbf{e}'_{p'}/\tau_{\mathrm{ph}}\right)
}.
\end{equation}
For supervised phrase-composition alignment, we match the predicted ROI-to-phrase distribution $\hat{q}_{i,k}$ to the softened target $q_{i,k}$, where both denote distributions over $p\in\mathcal{P}$:
\begin{equation}
\mathcal{L}_{\mathrm{comp}}
=
\frac{1}{|\Omega|}
\sum_{(i,k)\in\Omega}
D_{\mathrm{KL}}
\left(
q_{i,k}
\,\Vert\,
\hat{q}_{i,k}
\right),
\end{equation}
where $\Omega$ is the set of all image--region pairs.
This loss serves as a regularizer, encouraging ROI representations to remain consistent with both the localized annotations and their semantically related phrase variants. The overall TRPA objective is defined in Section~\ref{sec:training_inference}.

\begin{figure*}[t]
  \centering

  \begin{minipage}[b]{0.56\textwidth}
    \centering
    \includegraphics[width=\linewidth]{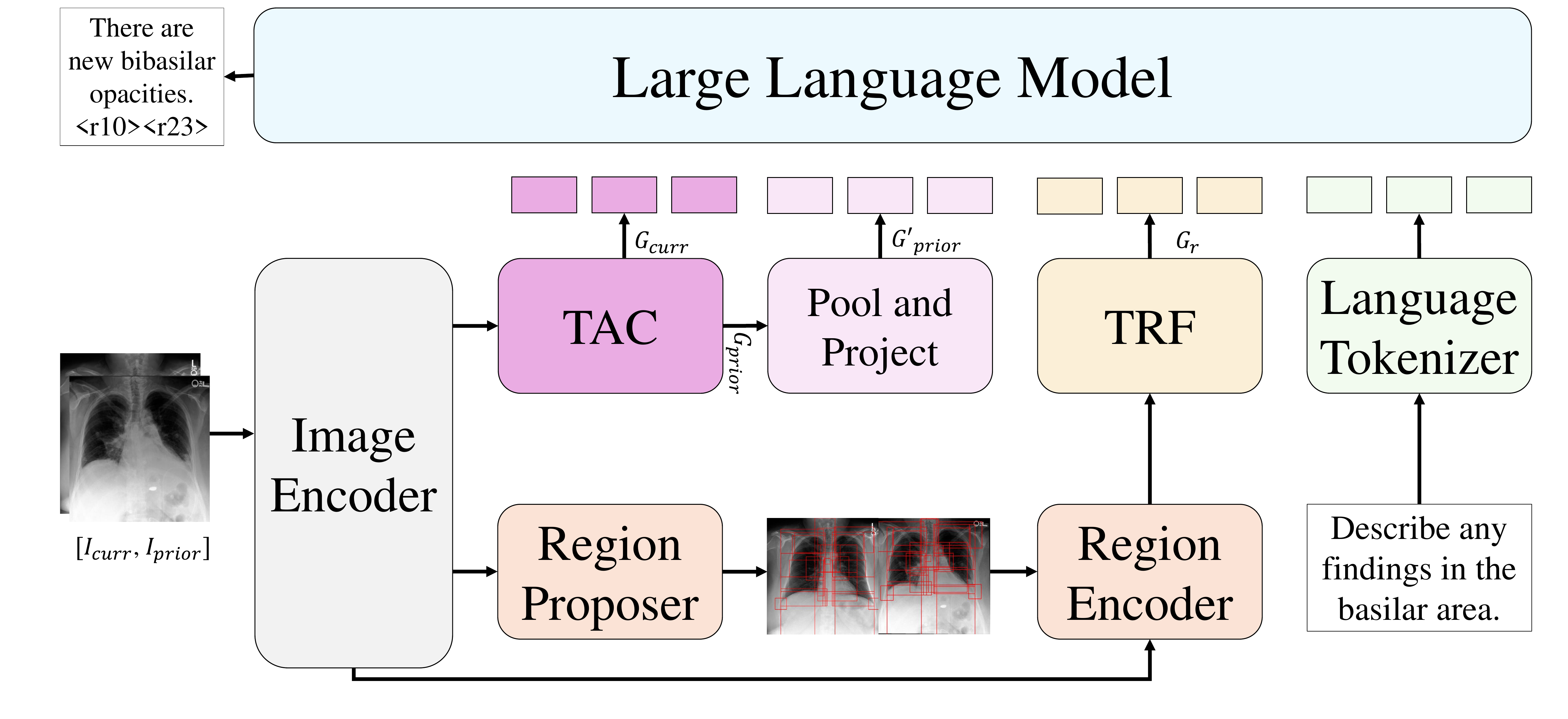}
    \textbf{(a)}
  \end{minipage}
  \hfill
  \begin{minipage}[b]{0.20\textwidth}
    \centering
    \includegraphics[width=\linewidth]{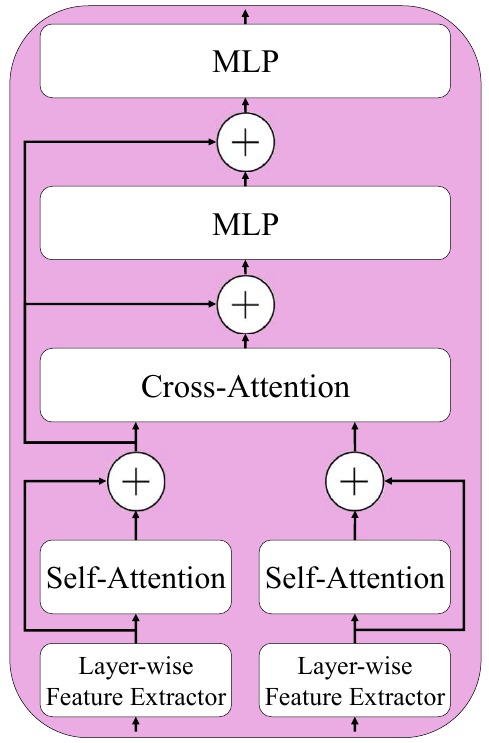}
    \textbf{(b)}
  \end{minipage}
  \hfill
  \begin{minipage}[b]{0.20\textwidth}
    \centering
    \includegraphics[width=\linewidth]{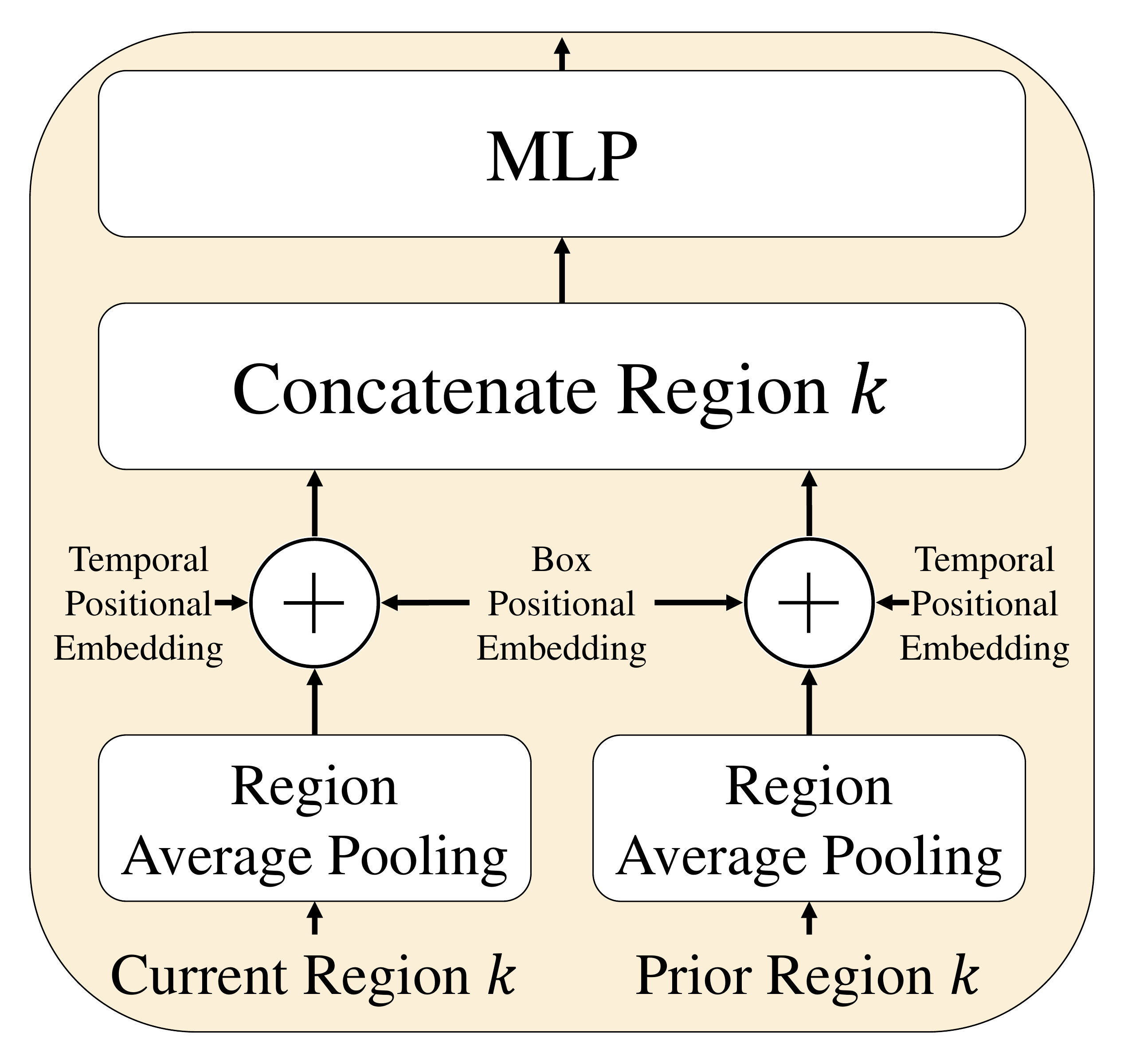}
    \textbf{(c)}
  \end{minipage}

  \vspace{0.5em}
  \caption{
~\textbf{CheXGround architecture.}
(a) Temporally aligned image tokens and anatomical ROI tokens are injected into the language model.
(b) The Temporal Alignment Connector (TAC), following Libra~\cite{zhang2025libra}, aligns current--prior image features.
(c) Temporal Region Fusion (TRF) fuses current--prior anatomical ROI representations.
}
  \label{fig:CheXGround_lm}
\end{figure*}
% =============================================================

\subsection{{\normalsize CheXGround: Region-Grounded Longitudinal Language Modeling}}
\label{sec:CheXGround_lm}

After TRPA pretraining, the language model receives two visual streams. 
The global stream carries temporally aligned image tokens from the current and prior radiographs, while the ROI stream carries fixed anatomical tokens from the detector and TRPA region encoder. 
Together, these streams provide both holistic temporal context and localized region-level evidence (see Figure~\ref{fig:CheXGround_lm}).

\noindent \textbf{Global image-token stream.}
Let \(\mathbf{F}_{\mathrm{curr}}\) and \(\mathbf{F}_{\mathrm{prior}}\) denote the RAD-DINO hidden features extracted from all layers for \(I_{\mathrm{curr}}\) and \(I_{\mathrm{prior}}\). We use a Temporal Alignment Connector (TAC), following Libra~\cite{zhang2025libra}, to align paired studies. TAC provides an effective mechanism for temporally aligning paired visual features by conditioning one study on prior (see Figure~\ref{fig:CheXGround_lm}(b)). However, in our longitudinal analysis setting, we found that a simple current-to-prior fusion operation is insufficient for modeling radiographic change (see ablations Table~\ref{tab:tac_ablation}). A one-way current-to-prior cross-attention can underrepresent the prior because current features remain dominant through self-attention and residual paths. We therefore construct a dual-TAC where current stream is conditioned on the prior, while the prior stream is self-conditioned through the same feature space.
\begin{equation}
\begin{aligned}
\mathcal{G}_{\mathrm{curr}} &= \mathrm{TAC}(\mathbf{F}_{\mathrm{curr}}, \mathbf{F}_{\mathrm{prior}}), \\
\mathcal{G}_{\mathrm{prior}} &= \mathrm{TAC}(\mathbf{F}_{\mathrm{prior}}, \mathbf{F}_{\mathrm{prior}}).
\end{aligned}
\end{equation}
For single-study inputs, the temporal path resolves to \(\mathrm{TAC}(\mathbf{F}_{\mathrm{curr}}, \mathbf{F}_{\mathrm{curr}})\). We retain the full current token sequence, but compress the prior sequence $\mathcal{G}_{\mathrm{prior}}$ to $\mathcal{G'}_{\mathrm{prior}}$ by concatenating neighboring tokens and applying channel-wise downsampling to reduce compute overhead, following~\cite{chen2310minigpt,ma2024groma}.

\noindent \textbf{Temporal Region Fusion (TRF).}
As illustrated in Figure~\ref{fig:CheXGround_lm}(c), the ROI stream supplies localized evidence from the pretrained TRPA region encoder. For each study $t$, we average-pool the extracted and enhanced regional features as
$r^{t}_{k}=\mathrm{AvgPool}(Z^{t}_{k})$.
We again add learnable box-position embeddings and sinusoidal time embeddings to obtain
$\mathbf{r}^{\mathrm{curr}}_{k}$ and $\mathbf{r}^{\mathrm{prior}}_{k}$ for the current and prior studies, respectively. The paired ROI token is
\begin{equation}
\mathbf{v}^{\mathrm{roi}}_{k}
=
\mathrm{g}_{\mathrm{roi}}
\left(
\mathbf{r}^{\mathrm{curr}}_{k}
\oplus
\mathbf{r}^{\mathrm{prior}}_{k}
\right),
\end{equation}
where $\oplus$ denotes channel-wise concatenation and $\mathrm{g}_{\mathrm{roi}}$ is an MLP that maps the paired local evidence into the language-model embedding space. Missing current or prior regions use a learned missing-region embedding before concatenation, preserving a fixed slot layout.

\noindent \textbf{Region-token interface.}
Each ROI embedding is paired with a discrete textual region identifier. For \(K\) anatomical slots, the region block is
\begin{equation}
    \mathcal{G}_{r}
    =
    [
    \langle r_1\rangle \langle \mathrm{region}\rangle_1,
    \ldots,
    \langle r_K\rangle \langle \mathrm{region}\rangle_K
    ],
\end{equation}
where the embedding of \(\langle \mathrm{region}\rangle_k\) is replaced by \(\mathbf{v}^{\mathrm{roi}}_k\). Thus, generated identifiers such as \(\langle r_7\rangle\) refer to the corresponding anatomical evidence token. During training, we randomly shuffle anatomical slots in \(\mathcal{G}_{r}\), which discourages reliance on a fixed order and encourages grounding through real semantic encoding.

\noindent \textbf{Multi-modal input sequence.}
Given an instruction \(x\), CheXGround forms
\begin{equation}
    \mathbf{M}
    =
    [
        \mathrm{Emb}(x),
        \langle \mathrm{curr}\rangle,
        \mathcal{G}_{\mathrm{curr}},
        \mathcal{G}_{r},
        \langle \mathrm{previm}\rangle,
        \mathcal{G}'_{\mathrm{prior}}
    ].
\end{equation}
The timestep markers $\langle \mathrm{curr}\rangle$ and $\langle \mathrm{previm}\rangle,$ help identify the two studies. We also aggregate the elapsed time between studies as natural-language context, such as ``prior image taken 150 days and 9 hours ago.'' The language model then generates the response autoregressively.
\begin{equation}
    p_{\theta}(y \mid x,I_{\mathrm{curr}},I_{\mathrm{prior}})
    =
    \prod_{n=1}^{N}
    p_{\theta}
    \left(
        y_n
        \mid
        y_{<n}, \mathbf{M}
    \right).
\end{equation}

%-------------------------------------------------------------------------
\subsection{Training}
\label{sec:training_inference}

We follow a staged curriculum comprising anatomical detector pretraining, temporal region--phrase alignment, vision--language interface alignment, and instruction tuning.

\noindent \textbf{Stage 0: anatomical detector pretraining.}
We first train an anatomy-aware detector from anatomical bounding box annotations. 
The detector follows a Deformable-DETR-style design~\cite{zhu2021deformable}, but replaces object queries with fixed anatomical queries assigned to predefined K anatomical regions. 
Since query identity specifies the target anatomy, no classification head is required. 
For each annotated image, anatomical query $k$ is directly supervised by the corresponding ground-truth box $\mathbf{b}^{*}_{t,k}$:
\begin{equation}
    \mathcal{L}_{\mathrm{det}}
    =
    \lambda_{1}
    \left\|
        \mathbf{b}_{t,k}
        -
        \mathbf{b}^{*}_{t,k}
    \right\|_{1}
    +
    \lambda_{\mathrm{giou}}
    \mathcal{L}_{\mathrm{GIoU}}
    \left(
        \mathbf{b}_{t,k},
        \mathbf{b}^{*}_{t,k}
    \right),
\end{equation}
where $\mathrm{GIoU}$ denotes generalized intersection over union, $\|\cdot\|_1$ is L1 loss and $\lambda_{1}$ and $\lambda_{\mathrm{giou}}$ are weights. 
During this stage, the image backbone is frozen and only detector-specific modules are optimized.

\noindent \textbf{Stage 1: temporal region--phrase alignment.}
With the anatomical detector and text encoder $\psi(\cdot)$ frozen, we train the spatial--temporal region encoder using the objectives in Section~\ref{sec:trpa}:
\begin{equation}
    \mathcal{L}_{\mathrm{TRPA}}
    =
    \lambda_{\mathrm{roi}}
    \mathcal{L}_{\mathrm{GLoRIA\text{-}ROI}}
    +
    \lambda_{\mathrm{comp}}
    \mathcal{L}_{\mathrm{comp}},
\end{equation}
where $\lambda_{\mathrm{roi}}$ and $\lambda_{\mathrm{comp}}$ are weighting coefficients.

\noindent \textbf{Stage 2: vision--language interface alignment.}
We then attach the pretrained region encoder to the language model and train only the vision--language interface, including temporal image connectors, token projectors, and region embeddings, using masked autoregressive language modeling:
\begin{equation}
    \mathcal{L}_{\mathrm{LM}}
    =
    -
    \sum_{n=1}^{N}
    \log
    p_{\theta}
    \left(
        y_n
        \mid
        y_{<n}, \mathbf{M}
    \right).
\end{equation}

\begin{table}[t]
\centering
\scriptsize
\setlength{\tabcolsep}{2.5pt}
\renewcommand{\arraystretch}{1.02}
\begin{tabular}{@{}p{0.29\linewidth}p{0.34\linewidth}rr@{}}
\hline
Task & Source & Train & Val \\
\hline
Report generation & MIMIC-CXR~\cite{johnson2019mimic, johnson2019mimic2} & 162,939 & 1,286 \\
CXR VQA & MIMIC-CXR-VQA~\cite{bae2023ehrxqa} & 284,077 & 61,974 \\
Difference VQA & Medical-Diff-VQA~\cite{hu2023medicaldiff} & 450,946 & 70,070 \\
Grounded / Referring VQA 
& \begin{tabular}[t]{@{}l@{}}
MIMIC-Ext-CXR-QBA~\cite{muller2026a}, \\
MIMIC-CXR-VQA~\cite{bae2023ehrxqa}
\end{tabular}
& 1,172,925 & 69,594 \\
Grounded Temporal VQA & MIMIC-Ext-CXR-QBA~\cite{muller2026a} & 129,280 & 1,034 \\
Anatomical Grounding & RadVLM~\cite{deperrois2025radvlm} & 72,196 & -- \\
\hline
\rowcolor{blue!7.5} Total & -- & 2,272,363 & 203,958 \\
\hline
\end{tabular}
\vspace{1.5em}
\caption{
Composition of the CheXGround vision--language training data across diverse radiology vision--language tasks.
}
\label{tab:dataset_composition}
\end{table}

\noindent \textbf{Stage 3: instruction tuning.}
Finally, we adapt the language model itself for grounded VQA, localized progression reasoning, and longitudinal report generation using $\mathcal{L}_{\mathrm{LM}}$. 

\section{Experiments}
\label{sec:experiments}

\noindent\textbf{Overview.}
We evaluate CheXGround on four primary task families.
Visual question answering (VQA), Grounded VQA, Longitudinal findings generation and Anatomy grounding. We also provide additional adaptations including abnormality grounding and temporal progression classification.

\begin{figure}[t]
    \centering
    \includegraphics[width=1.0\textwidth]{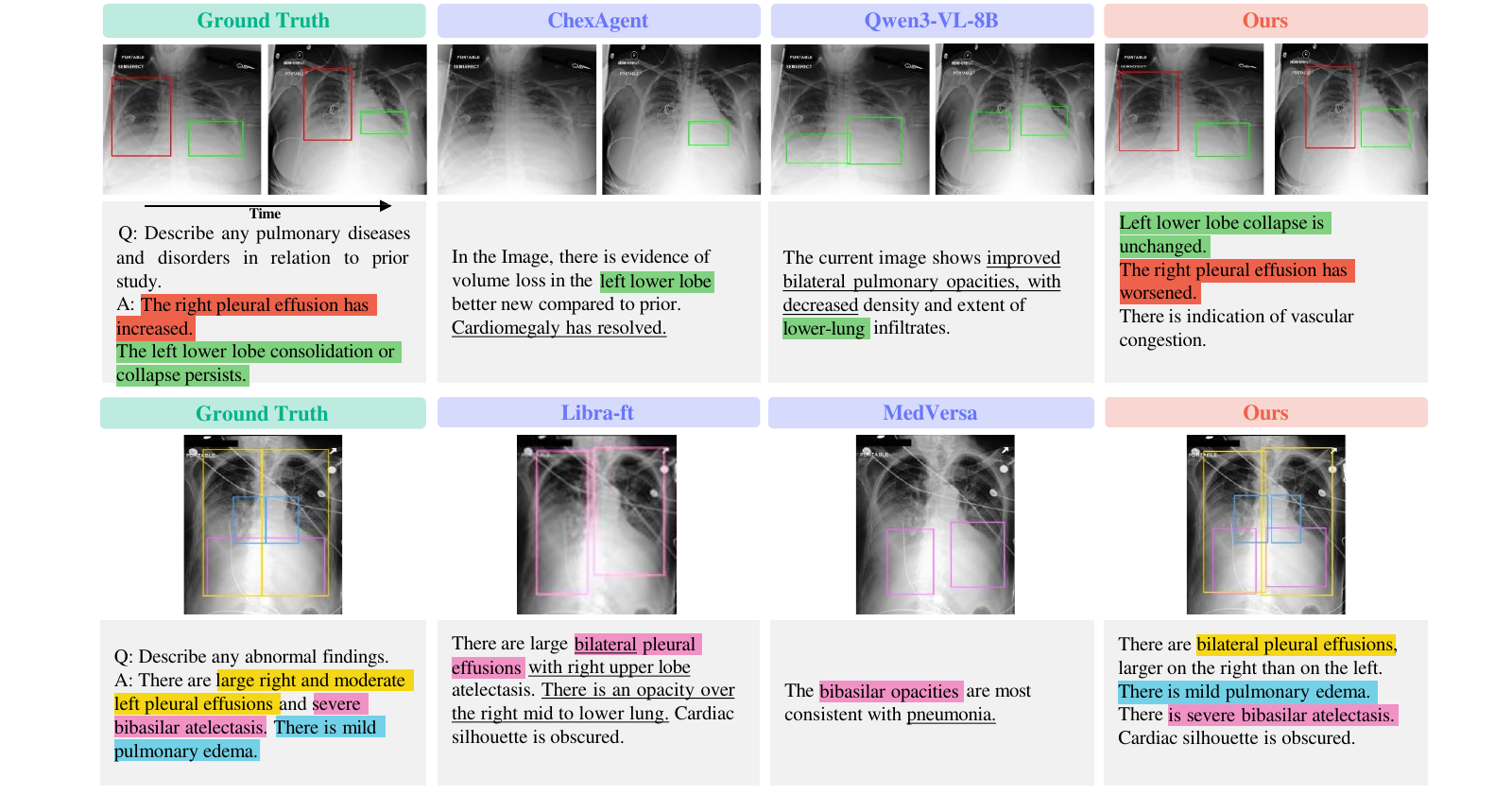}
    \captionsetup{skip=2pt}
    \caption{
Qualitative comparison of single-image and temporal grounding.
CheXGround shows more accurate anatomical localization and more faithful language than baselines, grounding findings in the correct regions while capturing longitudinal changes.
\underline{Underlined text} denotes inaccuracies. 
}
    \label{fig:grounding_results}
\end{figure}

\noindent\textbf{Implementation details.}
CheXGround uses RAD-DINO~\cite{perezgarcia2024raddino} as the radiology-tuned image encoder and a Meditron-7B language model initialized from the Libra multi-modal checkpoint~\cite{zhang2025libra,chen2023meditron}. 
For TRPA, the CXR-BERT text encoder is initialized from BioViL-T~\cite{bannur2023learning} and kept frozen to leverage its longitudinal radiology pretraining. 
We use $K=29$ anatomical regions following Chest ImaGenome~\cite{wu2021chest}. 
Training is organized into four stages: anatomical detector pretraining, TRPA alignment, visual-language interface pretraining, and LoRA instruction tuning. 
These stages use learning rates of $2{\times}10^{-4}$, $2{\times}10^{-4}$, $3{\times}10^{-5}$, and $2{\times}10^{-5}$ and effective batch sizes 48, 64, 24 and 32, respectively. 
All stages are trained on two NVIDIA A100 80GB GPUs using AdamW and cosine learning-rate scheduling.
The image encoder remains frozen throughout training. During both the visual-language stages, the base language model is also frozen, and only the multi-modal projectors, temporal modules, added tokens, and LoRA adapters are optimized. All newly introduced modules and TAC are trained from scratch.
The final instruction-tuning stage uses LoRA with rank $r=64$, scaling factor $\alpha=128$, and dropout $0.05$.

\begin{table}[t]
\begin{center}
\small
\resizebox{\linewidth}{!}{
\begin{tabular}{lccccc|ccccc}
\hline
 &  & \multicolumn{4}{c|}{Single-Study VQA} & \multicolumn{5}{c}{Longitudinal VQA} \\
\cline{3-6}\cline{7-11}
Model & Params
& BLEU-4 & ROUGE-L & METEOR & CheXbert-F1$_{\mathrm{micro/macro}}$
& BLEU-4 & ROUGE-L & METEOR & CheXbert-F1$_{\mathrm{micro/macro}}$ & $F1_{\mathrm{temp}}$ \\
\hline
MedGemma-1.5-4B
& 4B
& 0.1 & 1.7 & 4.8 & 25.5 / 13.5
& 8.4 & 3.8 & 8.5 & 23.1 / 15.5 & 43.7 \\

Libra
& 7B
& 6.4 & 24.4 & 18.7 & 60.3 / 24.4
& 6.4 & 40.4 & 26.4 & 54.5 / 24.1 & 70.7 \\

MedVersa
& 7B+
& 8.8 & 25.5 & 12.3 & 77.0 / 18.8
& 48.8 & 39.0 & 28.5 & 66.1 / \textbf{34.2} & 87.1 \\

CheXagent
& 8B
& 34.1 & 61.3 & 38.6 & 83.1 / 54.8
& 5.6 & 48.2 & 27.5 & 53.6 / 20.3 & 74.0 \\
\hline
\rowcolor{blue!7.5} CheXGround
& 7.4B
& \textbf{56.8} & \textbf{78.6} & \textbf{52.8} & \textbf{90.8 / 60.2}
& \textbf{50.6} & \textbf{75.9} & \textbf{52.8} & \textbf{71.7} / 34.0 & \textbf{97.7} \\
\hline
\end{tabular}}
\end{center}
\caption{Single-study and longitudinal visual question answering results.}
\label{tab:vqa_results}
\end{table}

\begin{table}[b]
\begin{center}
\small
\resizebox{\linewidth}{!}{%
\begin{tabular}{lccccccc}
\hline
Model 
& BLEU-4 & ROUGE-L & METEOR & CheXbert-F1$_{\mathrm{micro/macro}}$ & RadGraph-F1 & mIoU & Recall@0.5 \\
\hline

Qwen3-VL-8B~\cite{bai2025qwen3}$^\dagger$
& 2.1 & 15.1 & 21.5 & 42.9 / 28.3 & 20.7 & 11.2 & 20.6 \\

MedGemma-1.5-4B
& 3.7 & 16.5 & 21.2 & 39.9 / 27.2 & 11.6 & -- & -- \\

VividMed
& 3.5 & 21.1 & 19.1 & 21.0 / 26.8 & 7.4 & -- & -- \\

MedVersa$^\dagger$
& 9.8 & 24.1 & 19.6 & 33.0 / 16.8 & 12.6 & 17.8 & 24.3 \\

MAIRA-2
& 2.1 & 12.7 & 10.9 & 41.3 / 19.1 & 22.9 & 12.0 & 19.4 \\

CheXagent
& 3.9 & 23.5 & 19.7 & 62.0 / 29.9 & 24.8 & 13.6 & 13.5 \\

\hline
MedGemma-1.5-4B-ft
& 7.9 & 18.5 & 24.2 & 47.9 / 34.2 & 31.7 & 20.3 & 29.6 \\

Libra-ft 
& 10.2 & 25.1 & 20.1 & 55.0 / 35.8 & 33.2 & 29.8 & 32.3 \\

\hline
\rowcolor{blue!7.5} CheXGround
& \textbf{13.2} & \textbf{30.1} & \textbf{25.8} & \textbf{64.6 / 44.3} & \textbf{50.9} & \textbf{36.6} & \textbf{41.8} \\
\hline
\end{tabular}%
}
\end{center}
\caption{Grounded visual question answering on single-study images. $^\dagger$ denotes models evaluated with few-shot prompting.}
\label{tab:grounded_vqa_results}
\end{table}

\noindent\textbf{Datasets and grounded data construction.}
Stage~0 trains the anatomical detector with Chest ImaGenome region annotations from MIMIC-CXR frontal radiographs~\cite{johnson2019mimic, johnson2019mimic2, wu2021chest}. 
Stage~1 keeps the same dataset and uses MIMIC-CXR metadata to form temporal pair augmentations~\cite{johnson2019mimic}. 
Stages~2 and~3 use the diverse visual-language mixture in Table~\ref{tab:dataset_composition}. 

For grounded supervision, we extend MIMIC-CXR-VQA by interleaving answers with supporting anatomical annotations from the scene graph provided by the same dataset~\cite{bae2023ehrxqa}. From CXR-QBA~\cite{muller2026a}, we leverage the large-scale QA supervision with densely annotated answers making it especially useful for grounded and referring reasoning.
Since CXR-QBA annotations follow a region schema that differs from Chest ImaGenome, we use only the samples that can be mapped to our 29-region scheme, which provides comprehensive anatomical coverage. Temporal grounded samples use CXR-QBA's \emph{change} findings when available from its scene graph, giving CheXGround access to large-scale longitudinal supervision with multi-granular and dense annotations. All train/validation/test splits are patient-disjoint, source splits are assigned before task construction and inherited across all training stages.

\begin{table}[t]
\begin{center}
\small
\begin{minipage}{0.56\linewidth}
\centering
\resizebox{\linewidth}{!}{%
\begin{tabular}{lccccccc}
\hline
Model
& ROUGE-L & METEOR & CheXbert-F1$_{\mathrm{micro/macro}}$ & RadGraph-F1 & $F1_{\mathrm{temp}}$ & mIoU & Recall@0.5 \\
\hline
Qwen3-VL-8B$^\dagger$
& 15.9 & 12.7 & 43.5 / 26.5 & 17.6 & 19.9 & 5.4 & 15.8 \\

MedVersa$^\dagger$
& 9.7 & 11.1 & 32.2 / 18.3 & 15.9 & 24.2 & -- & -- \\

MedGemma-1.5-4B
& 10.8 & 20.3 & 44.5 / 32.2 & 19.4 & 20.1 & -- & -- \\

MAIRA-2
& 11.3 & 12.1 & 48.2 / 34.3 & 16.8 & 36.2 & -- & -- \\

CheXagent
& 17.6 & 16.7 & 48.1 / 36.3 & 21.3 & 32.1 & -- & -- \\
\hline
MedGemma-1.5-4B-ft
& 21.4 & 13.4 & 42.5 / 25.2 & 22.7 & 33.1 & 35.6 & 39.2 \\

Libra-ft
& 25.8 & 18.3 & 44.2 / 27.2 & 21.9 & 20.1 & 39.2 & 41.2 \\
\hline
\rowcolor{blue!7.5} CheXGround
& \textbf{31.8} & \textbf{20.3} & \textbf{51.5 / 38.2} & \textbf{24.2} & \textbf{40.1} & \textbf{49.1} & \textbf{56.3} \\
\hline
\end{tabular}%
}
\vspace{0.65em}
\caption{Temporal grounded visual question answering results. $^\dagger$ denotes models evaluated with few-shot prompting.}
\label{tab:temporal_grounded_vqa_results}
\end{minipage}
\hfill
\begin{minipage}{0.40\linewidth}
\centering
\resizebox{\linewidth}{!}{%
\begin{tabular}{lcccccc}
\hline
Model
& BLEU-4 & ROUGE-L & METEOR & CheXbert-F1$_{\mathrm{micro/macro}}$ & RadGraph-F1 & $F1_{\mathrm{temp}}$ \\
\hline
HERGen
& 7.8 & 19.8 & 9.0 & 41.2 / 27.2 & 19.7 & 14.2 \\

CheXagent
& 12.8 & 23.6 & 29.0 & 51.2 / 31.4 & 27.8 & 23.3 \\

MedGemma-1.5-4B
& 6.0 & 22.4 & 28.6 & 54.0 / 35.4 & 28.5 & 15.5 \\

MAIRA-2
& 22.1 & 38.3 & 36.4 & 57.0 / 43.4 & 36.9 & 40.1 \\

Libra
& 23.0 & 39.2 & 41.0 & 58.8 / 42.7 & 36.1 & 37.4 \\
\hline

\rowcolor{blue!7.5} CheXGround
& 24.4 & 41.6 & 41.4 & 59.4 / 43.4 & 36.7 & 40.2 \\

\rowcolor{blue!7.5} CheXGround-ft
& \textbf{27.4} & \textbf{43.6} & \textbf{45.4}
& \textbf{61.4 / 46.4} & \textbf{37.7} & \textbf{40.9} \\
\hline
\end{tabular}%
}
\vspace{0.65em}
\captionsetup{justification=centering,singlelinecheck=false}
\caption{Longitudinal findings generation results.}
\label{tab:longitudinal_report_generation_results}
\end{minipage}
\end{center}
\end{table}

\newcommand{\agimg}[1]{%
\includegraphics[width=\linewidth,height=1.42cm,keepaspectratio]{#1}%
}

\newcommand{\agcol}[1]{%
\begin{minipage}[b]{0.235\textwidth}
\centering
#1
\end{minipage}%
}

\newcommand{\agspace}{\hspace{-0.65em}}

\begin{figure*}[!t]
\centering

% -------- Row 1: SVC --------
\agcol{\agimg{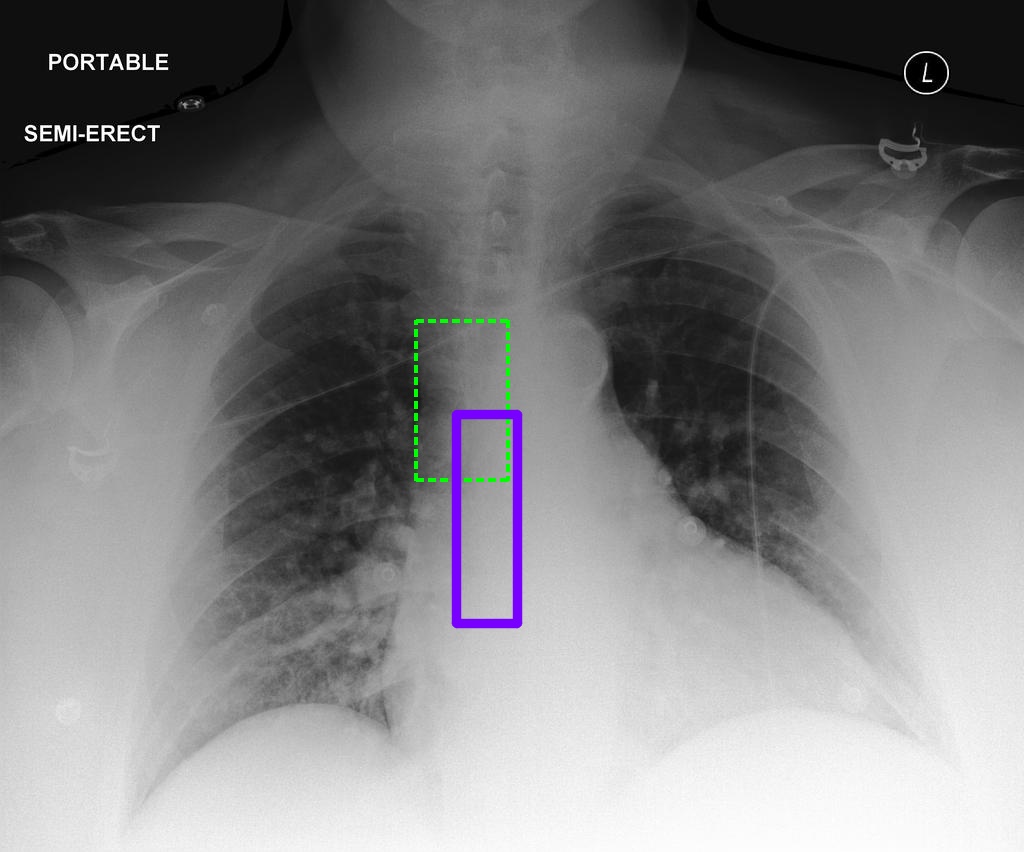}}
\agspace
\agcol{\agimg{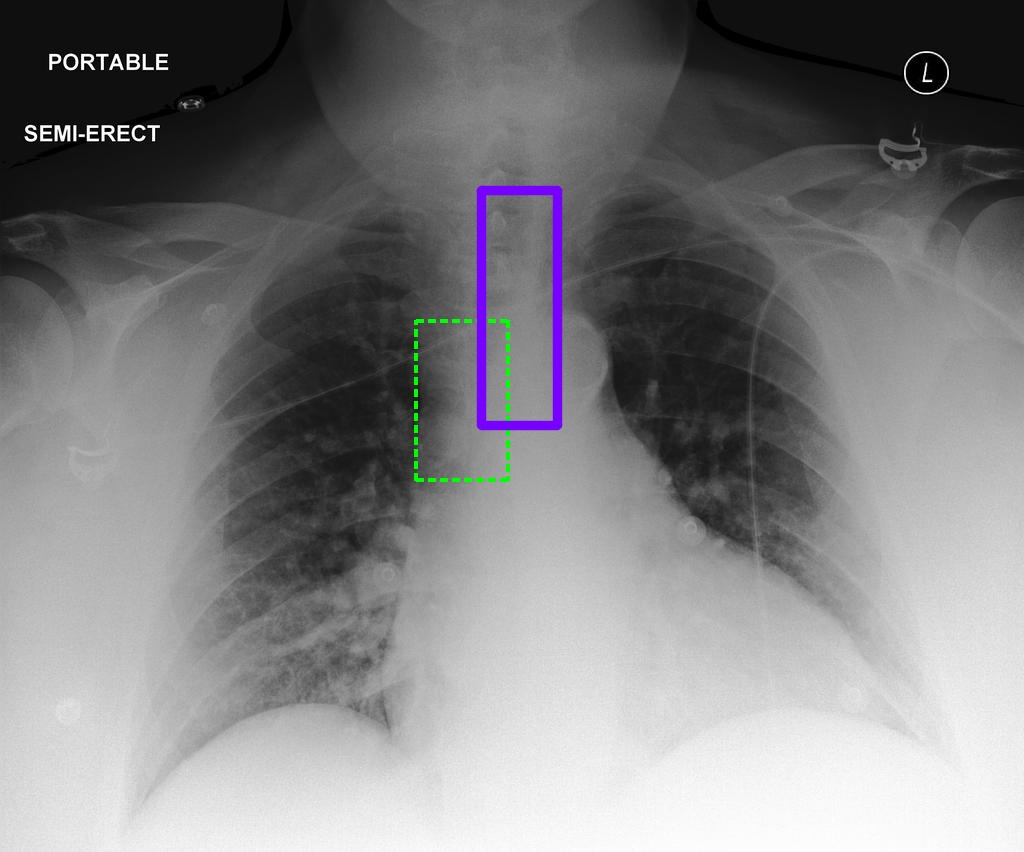}}
\agspace
\agcol{\agimg{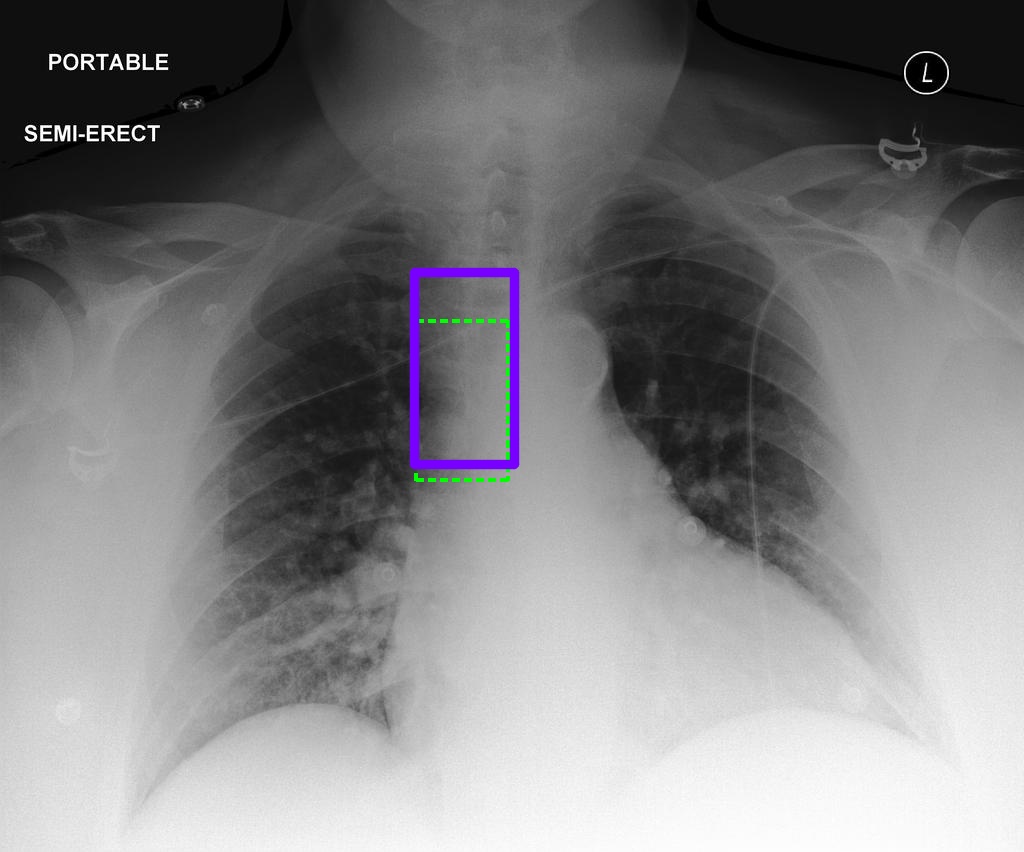}}
\agspace
\agcol{\agimg{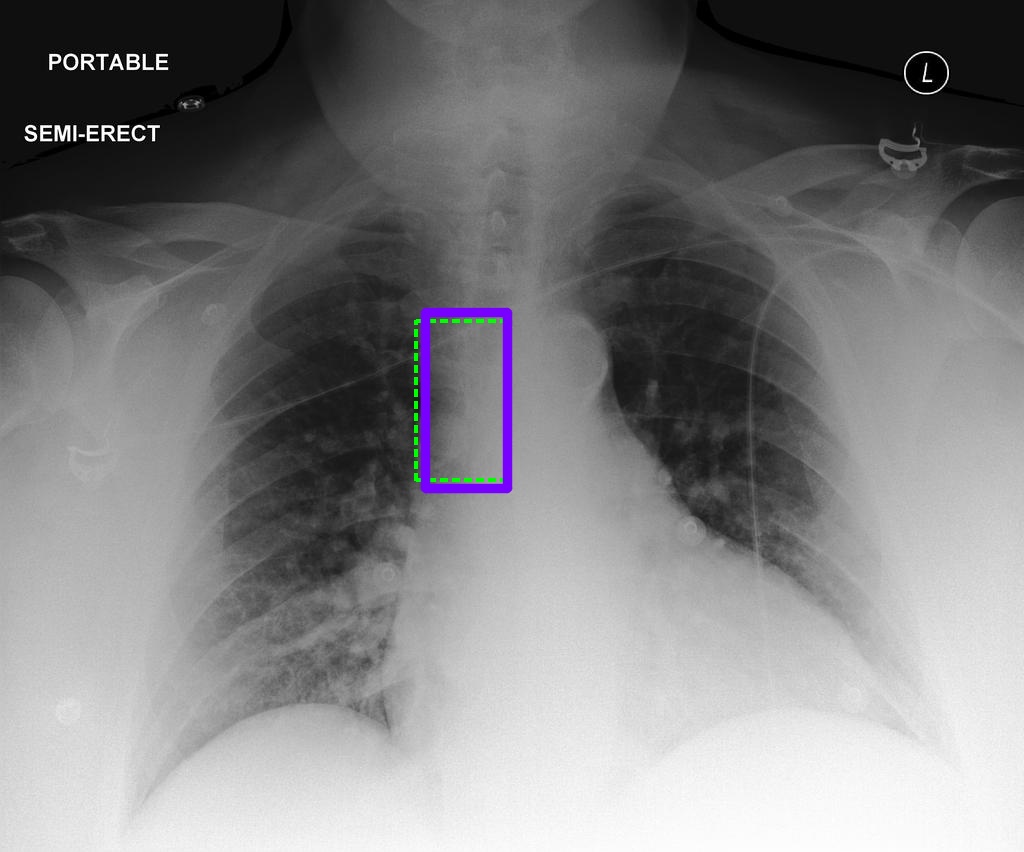}}

\vspace{-0.45em}

% -------- Row 2: Right costophrenic angle --------
\agcol{\agimg{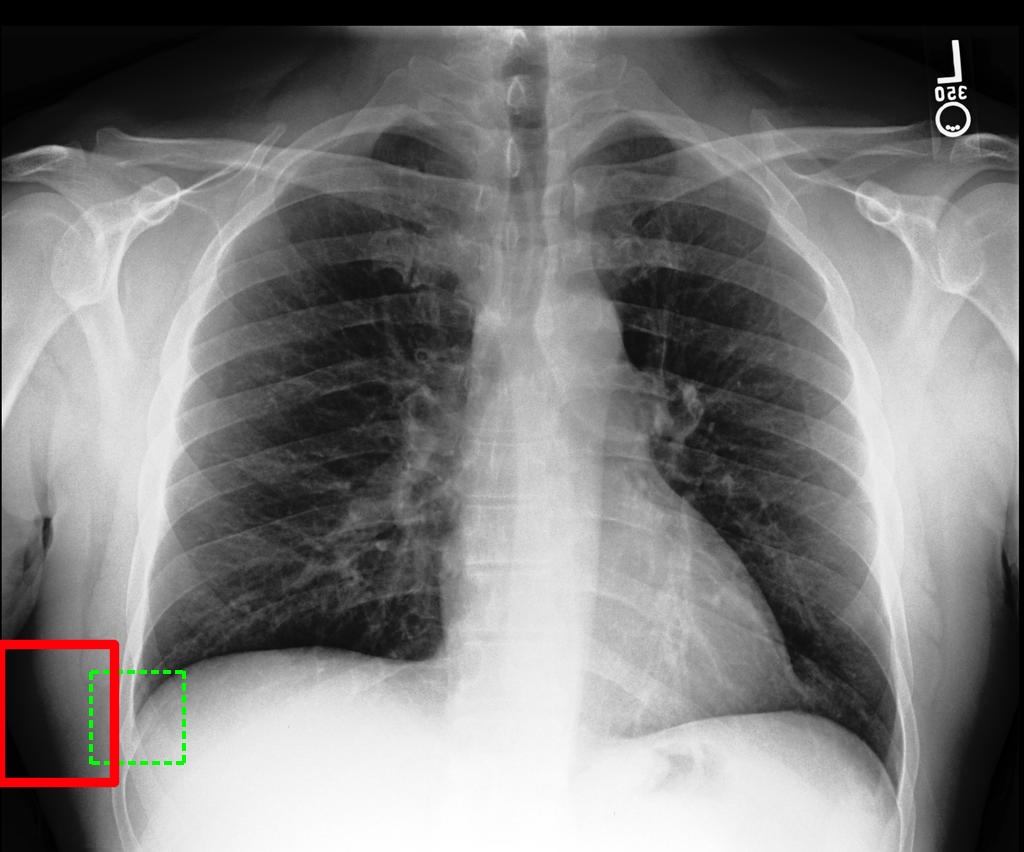}}
\agspace
\agcol{\agimg{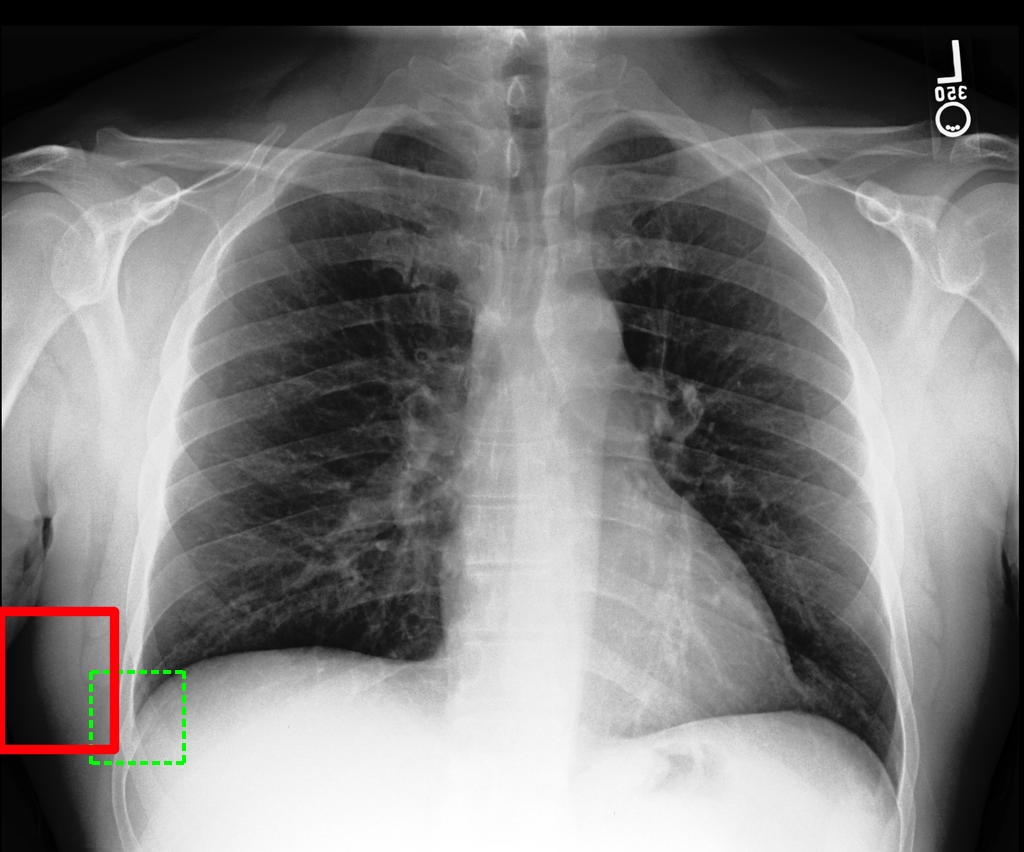}}
\agspace
\agcol{\agimg{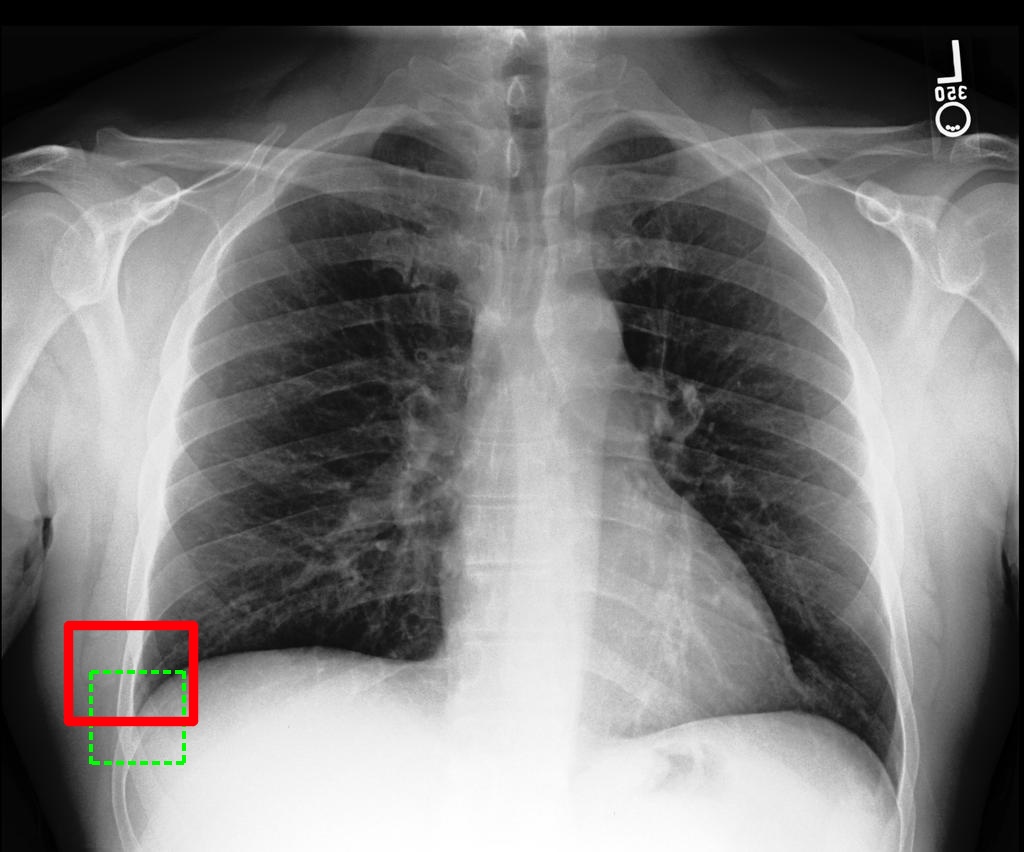}}
\agspace
\agcol{\agimg{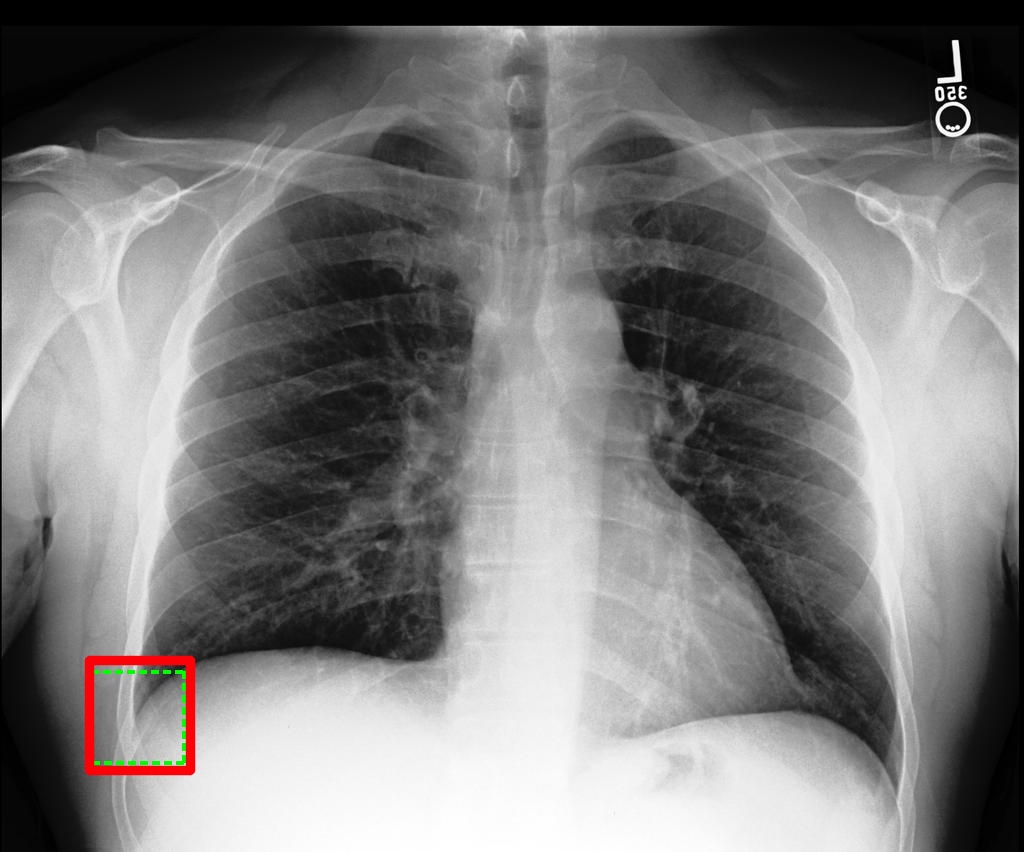}}

\vspace{-0.45em}

% -------- Row 3: Aortic arch --------
\agcol{\agimg{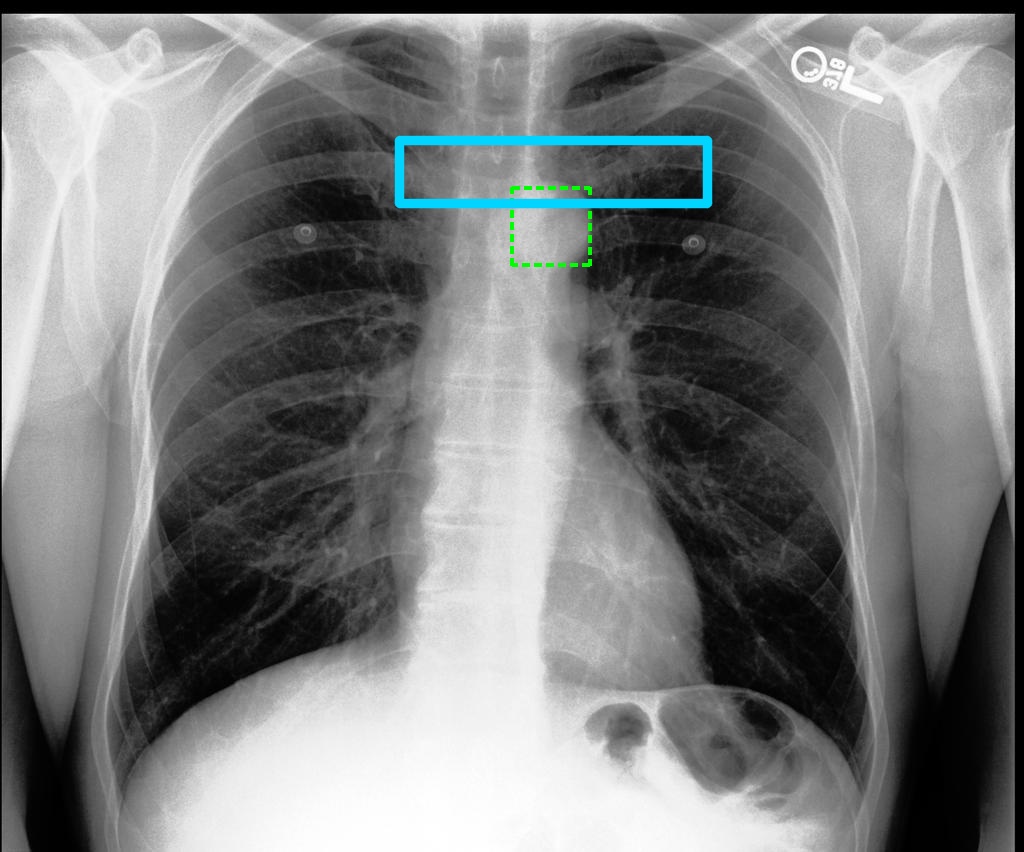}}
\agspace
\agcol{\agimg{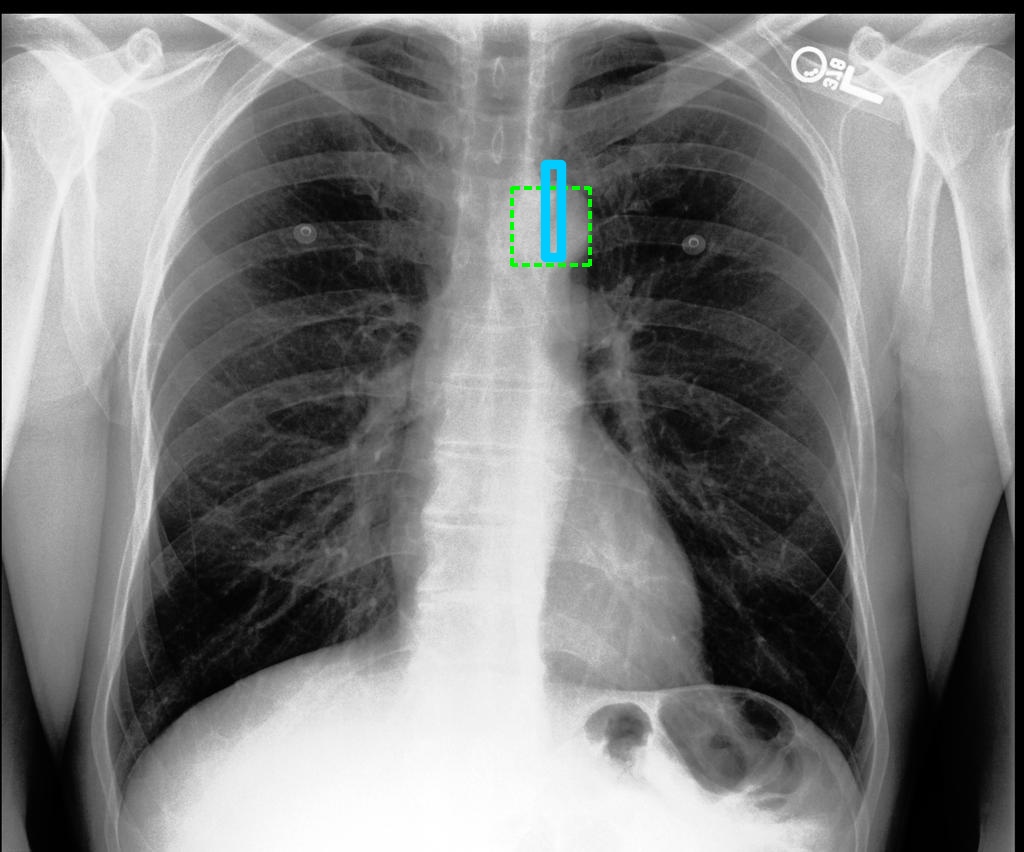}}
\agspace
\agcol{\agimg{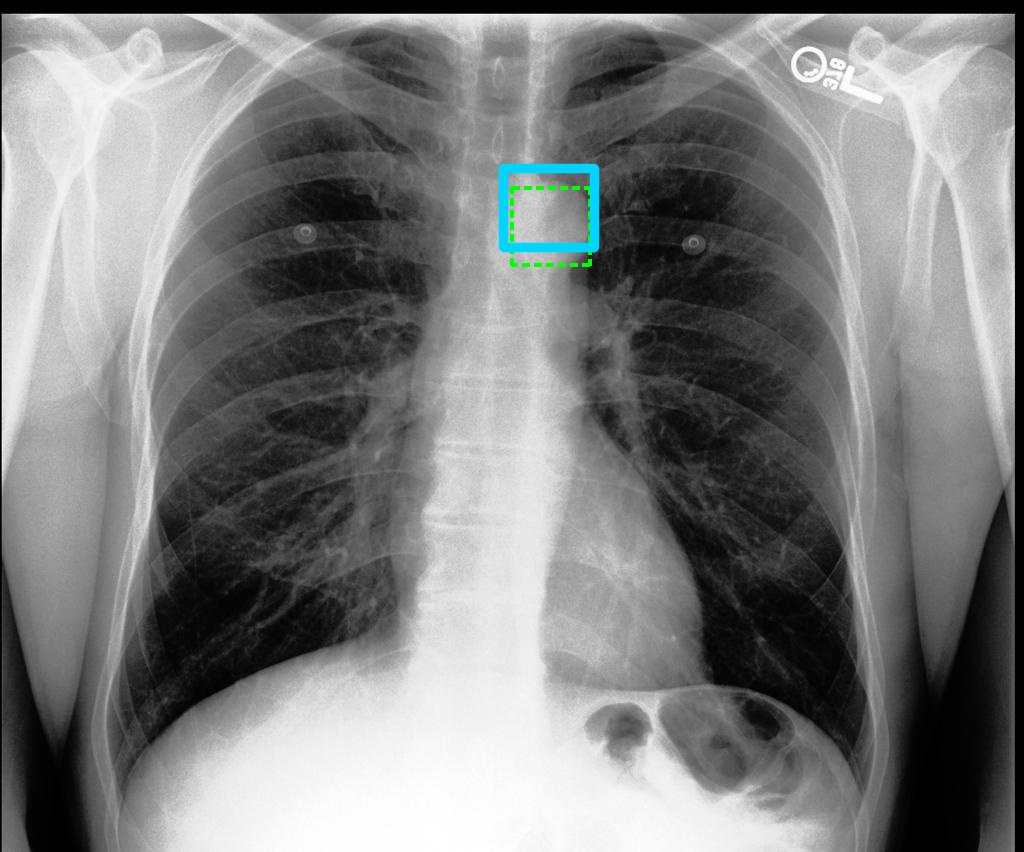}}
\agspace
\agcol{\agimg{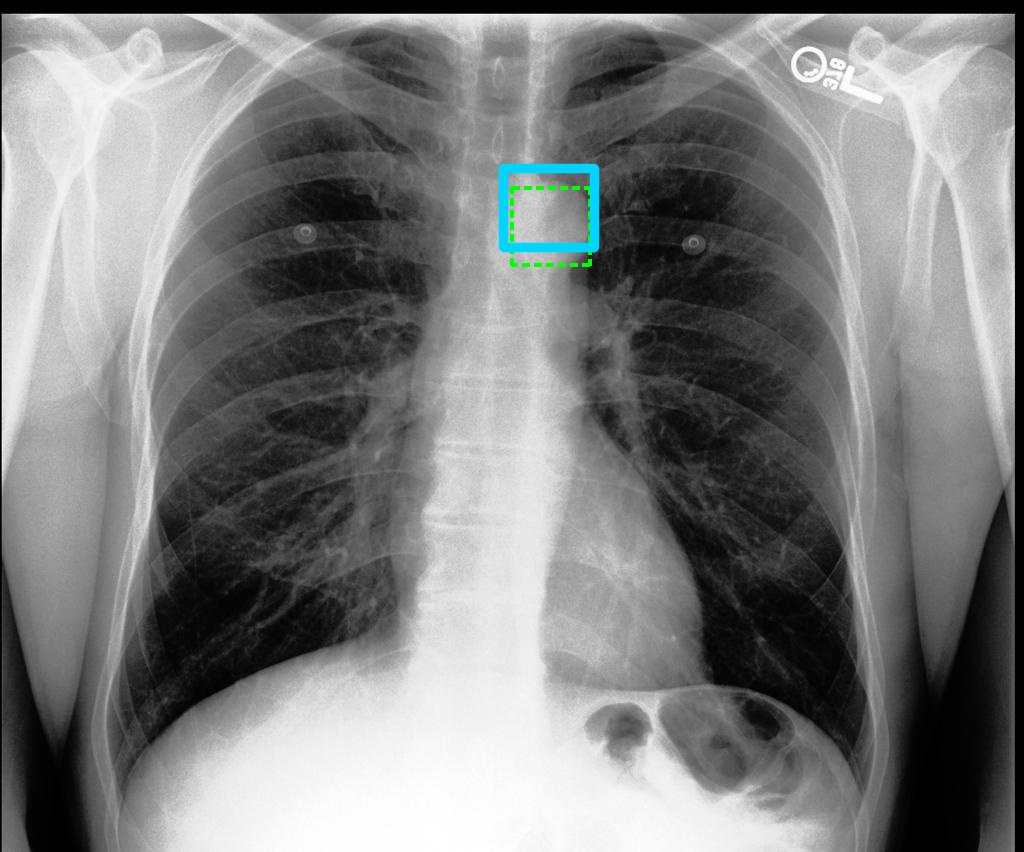}}

\vspace{0.15em}

% -------- Column labels --------
\makebox[0.235\textwidth]{\scriptsize Qwen3-VL-8B}
\agspace
\makebox[0.235\textwidth]{\scriptsize CheXagent}
\agspace
\makebox[0.235\textwidth]{\scriptsize RadVLM}
\agspace
\makebox[0.235\textwidth]{\scriptsize Ours}

\vspace{0.9em}

\caption{\textbf{Qualitative comparison on anatomy grounding.}
Rows show anatomical regions: SVC, right costophrenic angle, and aortic arch, respectively. Dashed green boxes represent ground truth; zoom in for clarity.}
\label{fig:anatomy_grounding}

\end{figure*}

\noindent\textbf{Evaluation protocol.}
Across generation tasks, we report BLEU, ROUGE-L~\cite{lin-2004-rouge}, METEOR~\cite{banerjee-lavie-2005-meteor}, RadGraph-F1~\cite{delbrouck-etal-2024-radgraph,jain2021radgraph}, and CheXbert-F1$_{\mathrm{micro/macro}}$ over all 14 CheXpert labels~\cite{smit2020chexbert}. Grounded VQA additionally reports mIoU and Recall@0.5, while anatomical grounding reports mIoU, Acc$_{50}$, and Acc$_{75}$, where Acc$_{\tau}$ is the percentage of predicted boxes with IoU at least $\tau$. For longitudinal VQA, temporal grounded VQA, and longitudinal findings generation, we also report temporal F1 following Libra~\cite{zhang2025libra}, evaluating lexical progression terminologies. All baselines are evaluated using the same test splits, preprocessing pipeline, and task-compatible prompts to ensure fair adaptation. For fair grounding evaluation, we further fine-tune MedGemma-1.5-4B~\cite{sellergren2026medgemma} and Libra on the same grounded instruction data for two epochs, denoted MedGemma-1.5-4B-ft and Libra-ft, and evaluate them under the same setting as CheXGround.
\vspace{0.5em}

\noindent\textbf{Visual and grounded question answering.}
Tables~\ref{tab:vqa_results}, \ref{tab:grounded_vqa_results}, and \ref{tab:temporal_grounded_vqa_results} evaluate single-study, longitudinal, and grounded VQA on MIMIC-CXR-VQA, Medical-Diff-VQA, and our held-out grounded splits~\cite{bae2023ehrxqa,hu2023medicaldiff,muller2026a}.
CheXGround achieves the strongest overall VQA performance.
High imbalance in lexical scores partly reflects the short-answer format: single/longitudinal answers average 1.80/5.86 tokens with medians 1/1, while canonical clinical strings can further increase overlap. Conversely, low BLEU can arise when answers use valid lexical variants that reduce exact n-gram overlap. We therefore complement lexical scores with clinical/semantic metrics which demonstrate a much fairer comparison.
For grounded VQA, CheXGround improves both answer quality and localization across single-study and temporal settings, including +9.9 mIoU and +15.1 Recall@0.5 over Libra-ft temporally.
Because \(F1_{\mathrm{temp}}\) does not verify whether predicted changes are grounded in the correct anatomy, we additionally report region-tag \(F1_{\mathrm{reg}}\) and coupled \(F1_{\mathrm{temp}\times\mathrm{reg}}\): CheXGround achieves 62.4/48.8 versus 43.7/37.7 for MedGemma-1.5-4B-ft and 46.8/28.1 for Libra-ft.

\begin{wraptable}{r}{0.48\textwidth}
    \centering
    \captionsetup{font=small,skip=3pt}
    \resizebox{\linewidth}{!}{%
    \begin{tabular}{lccc}
    \hline
    Model & mIoU & Acc$_{50}$ & Acc$_{75}$ \\
    \hline
    MedGemma-1.5-4B
    & 1.5 & 2.0 & 0.1 \\
    Qwen3-VL
    & 6.5 & 4.0 & 0.6 \\

    CheXagent
    & 18.9 & 17.0 & 9.3 \\

    RadVLM
    & 49.7 & 53.1 & 10.5 \\

    \hline
    \rowcolor{blue!7.5} CheXGround
    & \textbf{55.4} & \textbf{64.5} & \textbf{22.2} \\
    \hline
    \end{tabular}%
    }
    \captionsetup{justification=centering,singlelinecheck=false}
    \caption{Performance on the anatomical grounding task}
    \label{tab:anatomical_grounding_results}
\end{wraptable}

\noindent\textbf{Longitudinal findings generation.}
Table~\ref{tab:longitudinal_report_generation_results} evaluates free-form longitudinal findings generation solely on paired studies curated from the MIMIC-CXR test set~\cite{johnson2019mimic, johnson2019mimic2}.
Although report generation is not the primary focus of CheXGround, the general model remains competitive with dedicated findings-generation baselines and improves over Libra by +1.4 BLEU-4, +2.4 ROUGE-L, and +2.8 $F1_{\mathrm{temp}}$.
When tuned only for this task, CheXGround-ft further improves all metrics, reaching 43.57 ROUGE-L, 45.37 METEOR, 61.38/46.41 CheXbert-F1$_{\mathrm{micro/macro}}$, 37.7 RadGraph-F1, and 40.85 $F1_{\mathrm{temp}}$.
Taken together, these results indicate that CheXGround's temporal-grounded representation transfers to longitudinal findings generation. Unlike baselines optimized primarily for findings generation, our general model is trained across diverse tasks, which may introduce task-specific decoding noise in free-form reports.

\noindent\textbf{Anatomical grounding.}
Table~\ref{tab:anatomical_grounding_results} evaluates anatomical localization using coordinate-style queries curated from Chest ImaGenome~\cite{wu2021chest}.
CheXGround achieves the best performance across all metrics, with 55.4 mIoU, 64.5 Acc$_{50}$, and 22.2 Acc$_{75}$.

\noindent\textbf{Qualitative analysis.}
Figures~\ref{fig:anatomy_grounding} and~\ref{fig:grounding_results} show that CheXGround better links localized evidence to clinically faithful language. It grounds single-image findings and queried anatomical structures more tightly, while temporal examples show evidence localized across current and prior studies with correct progression direction. Baselines often miss temporal changes or produce absent-response and overextended grounding errors, whereas CheXGround can annotate regions across time through its temporally grounded architecture.

\noindent\textbf{Temporal Progression Classification and Abnormality grounding.}
We further evaluate transfer on MS-CXR-T~\cite{bannur2023learning} and abnormality-level grounding on VinDr-CXR-VQA and PadChest-GR~\cite{nguyen2025vindr,de_Castro_2025}. For MS-CXR-T, a disease query attends over the enhanced anatomical tokens for progression classification; fine-tuned TRPA improves macro-accuracy from 46.1 to 70.8, outperforming CoCa-CXR~\cite{chen2025cocacxrcontrastivecaptionerslearn}. For abnormality grounding, a lightweight LoRA adapter is trained over the frozen CheXGround components with coordinate-token box prediction as in MAIRA-2~\cite{bannur2024maira}, improving mIoU from 25.5 to 35.5 and outperforming RadVLM and MedVersa~\cite{deperrois2025radvlm,zhou2024medversa}. These experiments show that CheXGround's anatomy-centered temporal representations transfer to progression classification and abnormality localization. 

\noindent\textbf{Ablation studies.}
Table~\ref{tab:trpa_retrieval_ablation} evaluates Recall@1 ($\mathrm{R@1}$) TRPA using global image--report and ROI--phrase retrieval, with global scores computed from attention-pooled ROI representations. Direct ROI projection~\cite{ma2024groma}, global InfoNCE, and GLoRIA-style alignment improve retrieval to varying degrees, but lack explicit alignment to local anatomy. TRPA achieves the strongest local retrieval and downstream grounding by aligning fine-grained ROI tokens with clinical text, while removing temporal fusion $\mathcal{E}^{\mathrm{causal}}$ tests the importance of modeling temporal relationships. Table~\ref{tab:tac_ablation} ablates temporal alignment and anatomical evidence. We denote direct global token passing as N~\cite{bannur2024maira}, simple ROI projection as R~\cite{ma2024groma}, single-direction TAC as S~\cite{zhang2025libra}, spatio-temporal average pooling as A~\cite{qiu2024artemis}, and our dual-TAC and TRPA pretraining as D and P. Matched TRPA contrasts (P+R+N/R+N, P+R+S/R+S, P+R+D/R+D) improve language, clinical, grounding, and temporal metrics. The grounding effect is clearest in Acc$_{50}$, where TRPA adds +4.5/+3.5 over R+S/R+D, while P+R+D adds only +0.4 over P+R+S once TRPA is fixed. Thus, dual-TAC remains the larger temporal driver, while TRPA explains most localization gain and their combination gives the strongest overall performance.

\begin{table*}[t]
\centering
\small
\begin{minipage}[t]{0.49\textwidth}
\centering
\resizebox{\linewidth}{!}{
\begin{tabular}{lcccc}
\hline
Variant
& I$\rightarrow$R
& R$\rightarrow$I
& ROI$\rightarrow$Phr.
& Phr.$\rightarrow$ROI \\
\hline
ROI + projector (Groma) & 31.4 & 29.8 & 18.6 & 17.9 \\
global InfoNCE & 39.7 & 37.9 & 24.1 & 22.8 \\
GLoRIA & 41.2 & 39.4 & 31.5 & 30.2 \\
TRPA w/o $\mathcal{E}^{\mathrm{causal}}$ & 29.3 & 21.5 & 12.2 & 10.3 \\
TRPA w/o $\mathcal{L}_{\mathrm{comp}}$ & 45.9 & 44.1 & 39.6 & 38.2 \\
TRPA w/o $\mathcal{L}_{\mathrm{GLoRIA\mbox{-}ROI}}$ & 43.8 & 42.0 & 35.1 & 34.4 \\
\hline
\rowcolor{blue!7.5} TRPA & \textbf{49.6} & \textbf{47.8} & \textbf{46.3} & \textbf{44.9} \\
\hline
\end{tabular}}
\vspace{0.65em}
\caption{TRPA retrieval ($\mathrm{R@1}$) ablation.}
\label{tab:trpa_retrieval_ablation}
\end{minipage}
\hfill
\begin{minipage}[t]{0.49\textwidth}
\centering
{\setlength{\tabcolsep}{3.0pt}
\resizebox{\linewidth}{!}{
\begin{tabular}{lcccccc}
\hline
Setup
& ROUGE-L
& METEOR
& CheXbert-F1
& RadGraph-F1
& Acc$_{50}$
& $F1_t$ \\
\hline
R
& 45.8
& 29.9
& 42.7
& 19.2
& 43.2
& 24.7 \\

P+R (P-only)
& 49.0\gain{3.2}
& 30.0\gain{0.1}
& 44.2\gain{1.5}
& 23.9\gain{4.7}
& 45.1\gain{1.9}
& 27.8\gain{3.1} \\

N
& 53.2
& 35.8
& 58.7
& 25.8
& 56.2
& 41.5 \\

S
& 56.5
& 40.2
& 59.9
& 30.0
& 56.1
& 49.3 \\

D
& 58.3
& 42.6
& 61.8
& 30.5
& 57.5
& 53.2 \\

R+N
& 64.4
& 42.2
& 65.3
& 31.4
& 59.9
& 45.0 \\

P+R+N
& 70.2\gain{5.8}
& 46.7\gain{4.5}
& 68.9\gain{3.6}
& 36.2\gain{4.8}
& 60.3\gain{0.4}
& 51.1\gain{6.1} \\

R+A
& 64.5
& 43.5
& 66.0
& 32.8
& 61.0
& 53.9 \\

R+S
& 61.8
& 40.5
& 63.1
& 30.1
& 59.6
& 41.9 \\

P+R+S
& 67.8\gain{6.0}
& 46.9\gain{6.4}
& 67.4\gain{4.3}
& 35.6\gain{5.5}
& 64.1\gain{4.5}
& 42.9\gain{1.0} \\

R+D
& 69.1
& 48.2
& 68.1
& 37.5
& 61.0
& 63.0 \\
\hline
\rowcolor{blue!7.5}
\textbf{P+R+D}
& \textbf{75.9}\gain{\textbf{6.8}}
& \textbf{52.6}\gain{\textbf{4.4}}
& \textbf{71.7}\gain{\textbf{3.6}}
& \textbf{41.2}\gain{\textbf{3.7}}
& \textbf{64.5}\gain{\textbf{3.5}}
& \textbf{69.4}\gain{\textbf{6.4}} \\
\hline
\end{tabular}}}
\vspace{0.65em}
\caption{
Ablations on the effect of TAC and TRPA, on CheXGround. \gain{} denotes gain of P (TRPA) over the respective baseline.} 
\label{tab:tac_ablation}
\end{minipage}
\end{table*}

\section{Conclusion}
\label{sec:conclusion}

We introduced CheXGround, a region-grounded longitudinal chest X-ray language model that represents current and prior studies through corresponding anatomical regions. 
CheXGround combines global temporal image context with phrase-aligned temporal ROI tokens, enabling localized anatomical evidence to support language generation over paired studies. 
Through Temporal Region--Phrase Alignment pretraining, these region representations are aligned with localized clinical phrases across time. 
Across single-study VQA, longitudinal VQA, longitudinal findings generation, grounded VQA, temporal grounded VQA, and anatomical grounding, CheXGround consistently improves clinical language quality, temporal reasoning, and localization accuracy over recent baselines. 
These findings highlight anatomy-level temporal grounding as an effective framework for longitudinal radiology language modeling.

\section*{Acknowledgments}
This research was funded by Khalifa University of Science and Technology through the Faculty Start-Ups under Project ID: KU-INT-FSU-2005-8474000775.

\bibliography{egbib}

\end{document}